\documentclass{article}

\PassOptionsToPackage{numbers, compress}{natbib}

\usepackage[preprint]{neurips_2026}

\usepackage[utf8]{inputenc} 
\usepackage[T1]{fontenc}    
\usepackage{hyperref}       
\usepackage{url}            
\usepackage{booktabs}       
\usepackage{amsfonts}       
\usepackage{nicefrac}       
\usepackage{microtype}      
\usepackage{xcolor}         
\usepackage{placeins}

\usepackage{tabularx}
\usepackage{graphicx}
\usepackage{array}
\usepackage{makecell}
\usepackage{colortbl}
\usepackage{caption}
\usepackage{subcaption}
\usepackage{enumitem}
\usepackage{subcaption}  

\usepackage{adjustbox}
\usepackage{rotating}  
\usepackage{booktabs, multirow, makecell, colortbl}

\definecolor{oursrow}{RGB}{212,237,218}    
\definecolor{hdrblue}{RGB}{214,228,247}    
\definecolor{checkgreen}{RGB}{26,107,42}   
\definecolor{crossred}{RGB}{179,0,0}       
\definecolor{partgray}{RGB}{100,100,100}   

\newcommand{\cmark}{\textcolor{checkgreen}{\textbf{\checkmark}}}
\newcommand{\xmark}{\textcolor{crossred}{\textbf{$\times$}}}
\newcommand{\pmark}{\textcolor{partgray}{\textbf{\texttildelow}}}
\newcommand{\husic}{\textsc{husic}}
\newcommand{\ourname}{Urban-ImageNet}

\title{Urban-ImageNet: A large-scale multi-modal dataset and evaluation framework for urban space perception}

\author{%
	\textbf{Yiwei Ou}$^{1\,\dagger}$\thanks{%
		*Equal contributions.\quad
		$^{\dagger}$Corresponding author, email:
		\texttt{you661@aucklanduni.ac.nz}}\ \quad
	\textbf{Chung Ching Cheung}$^{2\,*}$ \quad
	\textbf{Jun Yang Ang}$^{3\,*}$ \quad
	\textbf{Xiaobin Ren}$^{1\,*}$
	\\[0.4em] 
	\textbf{Ronggui Sun}$^{1\,*}$ \quad
	\textbf{Guansong Gao}$^{1}$ \quad
	\textbf{Kaiqi Zhao}$^{4}$ \quad
	\textbf{Manfredo Manfredini}$^{1}$
	\\[0.8em]
	$^{1}$University of Auckland \quad
	$^{2}$University of Pennsylvania \quad
	$^{3}$Stanford University
	\\[0.4em]
	$^{4}$Harbin Institute of Technology, Shenzhen
}

\begin{document}

\maketitle

\begin{abstract}
We present Urban-ImageNet, a large-scale multi-modal dataset and evaluation benchmark for urban space perception from user-generated social media imagery. The corpus contains over \textbf{2~Million} public social media images and paired textual posts collected from Weibo across \textbf{61 urban sites} in \textbf{24~Chinese cities} across 2019–2025, with controlled benchmark subsets at 1K, 10K, and 100K scale and a full 2M corpus for large-scale training and evaluation. Urban-ImageNet is organized by HUSIC, a \textbf{H}ierarchical \textbf{U}rban
\textbf{S}pace \textbf{I}mage \textbf{C}lassification framework that defines a 10-class taxonomy grounded in urban theory. The taxonomy is designed to distinguish activated and non-activated public spaces, exterior and interior urban environments, accommodation spaces, consumption content, portraits, and non-spatial social-media content. Rather than treating urban imagery as generic scene data, Urban-ImageNet evaluates whether machine perception models can capture spatial, social, and functional distinctions that are central to urban studies. The benchmark supports three tasks within one standardized library: \textbf{(T1)} urban scene semantic classification, \textbf{(T2)} cross-modal image–text retrieval, and \textbf{(T3)} instance segmentation. Our experiments evaluate representative vision, vision-language, and segmentation models, revealing strong performance on supervised scene classification but more challenging behavior in cross-modal retrieval and instance-level urban object segmentation. A multi-scale study further examines how model performance changes as balanced training data increases from 1K, 10K to 100K images. Urban-ImageNet provides a unified, theory-grounded, multi-city benchmark for evaluating how AI systems perceive and interpret contemporary urban spaces across modalities, scales, and task formulations. All data, models, and code are publicly released.%
		\footnote{Data: \url{huggingface.co/datasets/Yiwei-Ou/Urban-ImageNet};
			Benchmark: \url{github.com/yiasun/dataset-2}}

\end{abstract}

\section{Introduction}
\label{sec:intro}

Understanding how people perceive and interact with urban spaces is a
fundamental challenge in urban planning, architecture, and smart-city
research. Traditional approaches---field surveys and in-situ
observation---are costly, spatially constrained, and difficult to scale
\citep{gehl2011,whyte1980}. The rise of social media has created an
unprecedented opportunity: billions of geotagged user-generated images
now document, at large scale, how people selectively experience and
represent urban environments \citep{hochman2013,boy2017}. 
Unlike systematically sampled imagery, these user-generated images constitute
\emph{revealed preference data}: users actively choose what to photograph and
share, making image frequency itself a quantitative proxy for spatial
attractiveness and place perception.
Despite this opportunity, no existing dataset is designed to harness social
media imagery for the study of urban space perception---leaving critical
gaps between the richness of available data and the tools researchers need to
exploit it.

\textbf{Gap 1: No multi-task urban space perception dataset.}
Advancing urban studies requires three complementary capabilities:
\textbf{(T1)}~scene-level semantic classification,
\textbf{(T2)}~cross-modal image--text retrieval, and
\textbf{(T3)}~instance-level segmentation.
Existing resources address each task in isolation---Places365
\citep{zhou2018places} and SUN \citep{xiao2010sun} for T1; MS-COCO
\citep{lin2014coco}, Flickr30K \citep{young2014flickr30k}, and LAION-5B
\citep{schuhmann2022laion} for T2; Cityscapes \citep{cordts2016cityscapes}
and LVIS \citep{gupta2019lvis} for T3---forcing researchers to stitch
together incompatible label systems and blocking
the development of unified urban scene understanding models.

\textbf{Gap~2: No standardised benchmarking library exists.}
Existing collections provide no shared tooling:
researchers must independently implement data loaders, fine-tuning pipelines,
and evaluation scripts for each task--dataset combination.
No library supports T1, T2, and T3 simultaneously in a single framework, nor provides standardised adapters for comparison
against established general-purpose benchmarks, making cross-study
reproducibility and comparability structurally limited.

\textbf{Gap~3: Existing urban scene taxonomies lack theoretical grounding.}
Benchmarks such as Places365 and SUN derive categories from web-image
frequency rather than conceptual distinctions relevant to urban research, while
urban perception datasets such as Place~Pulse~2.0 \citep{dubey2016deeplearning}
and MMS-VPR \citep{ou2025mmsvpr} introduce urban relevance but restrict
coverage to exterior street-level views.
Without labels grounded in Lefebvre's conceived versus lived space
\citep{lefebvre1991}, Gehl's social activity framework \citep{gehl2011}, or
Newman's public-to-private spatial gradient \citep{newman1972}, learned
representations cannot distinguish a socially activated public space from an
architecturally identical but empty one, nor support downstream tasks such as
spatial vitality mapping or retail environment assessment.

\begin{figure*}[t]
		\centering
		\includegraphics[width=\textwidth]{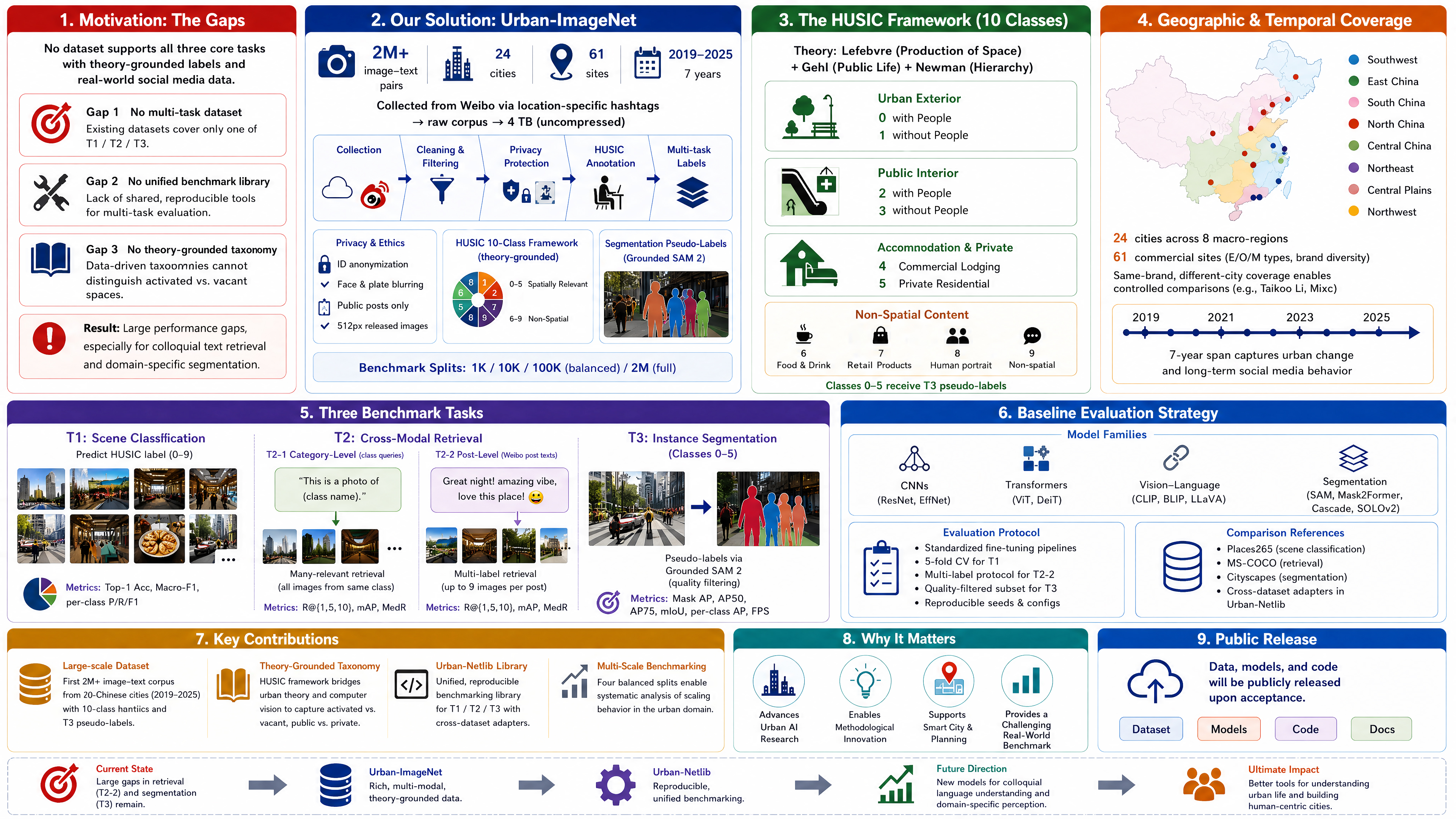}
		\caption{Urban-ImageNet framework: addressing current limitations in urban perception evaluation.}
		\label{fig:framework}
	\end{figure*}

To address these gaps, we present \textbf{\ourname{}}, with three
contributions (Figure~\ref{fig:framework}):

\begin{enumerate}[leftmargin=1.8em,itemsep=2pt,topsep=4pt]
  \item \textbf{\ourname{} Dataset.}
        Over \textbf{2M} image--text pairs from 24 Chinese cities, 61
        commercial sites, and seven years (2019--2025), with four benchmark
        splits (1K\,/\,10K\,/\,100K strictly balanced; 2M full corpus) and
        manual annotations for all three tasks. It simultaneously provides
        UGC origin, theory-grounded labels, multi-modal pairing, multi-city
        longitudinal coverage, and domain-specific instance segmentation
        pseudo-labels for urban space perception research.
\item \textbf{\ourname{}-lib.}
      A benchmarking library supporting T1, T2, and T3 within
      a single unified framework, supporting cross-dataset
      comparison against Places365, MS-COCO, and Cityscapes.
  \item \textbf{\husic{} Framework.}
        A 10-class urban space taxonomy grounded in Lefebvre
        \citep{lefebvre1991}, Gehl \citep{gehl2011}, and Newman
        \citep{newman1972}, capturing theoretically motivated distinctions
        between activated and non-activated spaces, publicly integrated and
        private interiors, and exterior and indoor commercial
        environments---distinctions absent from all existing large-scale
        vision benchmarks.
\end{enumerate}

\section{Related Work}
\label{sec:related}

\textbf{Urban Scene Classification.} SUN~\citep{xiao2010sun}, MIT Indoor~\citep{quattoni2009indoor}, and
Places365~\citep{zhou2018places} are the dominant scene classification
benchmarks, all constructed by keyword-based web crawling with no theoretical
grounding or textual metadata.
A Places365 model identifies a ``shopping mall'' but cannot distinguish a
socially activated commercial interior from an architecturally identical
empty space---a distinction central to public space analysis
\citep{gehl2011,whyte1980}.

\textbf{Image--Text Retrieval and Urban Perception.}
Flickr30K~\citep{young2014flickr30k} and MS-COCO Captions~\citep{lin2014coco}
provide objective third-person captions; LAION-5B~\citep{schuhmann2022laion}
demonstrates the power of web-harvested alt-text at billion scale.
None captures authentic first-person spatial narratives of social media.
Urban perception datasets including Place Pulse 2.0
\citep{dubey2016deeplearning}, MMS-VPR~\citep{ou2025mmsvpr}, and UrbanFeel
\citep{he2025urbanfeel} focus on exterior street-level imagery, provide no
textual modality, and lack multi-task annotation.

\textbf{Instance Segmentation.}
Cityscapes~\citep{cordts2016cityscapes}, ADE20K~\citep{zhou2019ade20k},
LVIS~\citep{gupta2019lvis}, and Mapillary Vistas~\citep{neuhold2017mapillary}
cover outdoor driving and general scenes but apply no domain-specific
vocabulary tailored to commercial spaces---the escalators, retail shelves,
display cases, hotel beds, and food presentations that define the majority of
\ourname{}'s images.

\textbf{Scaling Behaviour.}
ImageNet~\citep{deng2009imagenet} established scale as a performance driver;
GPT-3~\citep{brown2020gpt3} and scaling laws~\citep{kaplan2020scaling} showed
predictable growth; LAION-5B~\citep{schuhmann2022laion} demonstrated
billion-scale vision--language benefits.
\ourname{}'s multi-tier release enables scaling study
in the urban research domain.

\textbf{Comparison with Existing Datasets.} Table~\ref{tab:comparison} compares \ourname{} across task support, data
properties, and research applicability.
\ourname{} is the only dataset simultaneously providing UGC origin,
theory-grounded labels, multi-modal image--text pairing, multi-city
longitudinal coverage, and per-class instance segmentation labels.

\begin{table*}[h]
\centering
\caption{Unified comparison of \ourname{} with existing datasets. (T1\,=\,Scene Classification,
T2\,=\,Cross-Modal Retrieval, T3\,=\,Instance Segmentation), \cmark~supported; \xmark~not supported; \pmark~partially.
}
\label{tab:comparison}
\setlength{\tabcolsep}{3.0pt}
\renewcommand{\arraystretch}{1.22}
\resizebox{\textwidth}{!}{%
\begin{tabular}{
  l
  r
  ccc
  ccc
  cccc
}
\toprule
& &
\multicolumn{3}{c}{\textbf{Task Support}} &
\multicolumn{3}{c}{\textbf{Data Properties}} &
\multicolumn{4}{c}{\textbf{Research Applicability}} \\
\cmidrule(lr){3-5}
\cmidrule(lr){6-8}
\cmidrule(lr){9-12}
\textbf{Dataset} &
\makecell[r]{\textbf{Scale}\\(images)} &
\makecell{\textbf{T1}\\Cls.} &
\makecell{\textbf{T2}\\Ret.} &
\makecell{\textbf{T3}\\Seg.} &
\makecell{UGC /\\Social} &
\makecell{Multi-\\city} &
\makecell{Temporal\\($\ge$3\,yr)} &
\makecell{Theory-\\grnd.} &
\makecell{Urban\\comm.} &
\makecell{Asian\\cities} &
\makecell{Per-cls\\seg.} \\
\midrule
\multicolumn{12}{l}{\textit{Scene classification datasets}} \\
\quad Places365~{\scriptsize\citep{zhou2018places}}
    & 1.8M   & \cmark & \xmark & \xmark & \xmark & \cmark & \xmark & \xmark & \xmark & \xmark & \xmark \\
\quad SUN~{\scriptsize\citep{xiao2010sun}}
    & 131K   & \cmark & \xmark & \xmark & \xmark & \xmark & \xmark & \xmark & \xmark & \xmark & \xmark \\
\quad MIT Indoor~{\scriptsize\citep{quattoni2009indoor}}
    & 15.6K  & \cmark & \xmark & \xmark & \xmark & \xmark & \xmark & \xmark & \cmark & \xmark & \xmark \\
\midrule
\multicolumn{12}{l}{\textit{Segmentation datasets}} \\
\quad Cityscapes~{\scriptsize\citep{cordts2016cityscapes}}
    & 25K    & \xmark & \xmark & \cmark & \xmark & \cmark & \xmark & \xmark & \xmark & \xmark & \xmark \\
\quad ADE20K~{\scriptsize\citep{zhou2019ade20k}}
    & 27.5K  & \pmark & \xmark & \cmark & \xmark & \xmark & \xmark & \xmark & \xmark & \xmark & \xmark \\
\quad LVIS~{\scriptsize\citep{gupta2019lvis}}
    & 164K   & \xmark & \xmark & \cmark & \xmark & \xmark & \xmark & \xmark & \xmark & \xmark & \xmark \\
\quad Mapillary Vistas~{\scriptsize\citep{neuhold2017mapillary}}
    & 25K    & \xmark & \xmark & \cmark & \xmark & \cmark & \xmark & \xmark & \xmark & \pmark & \xmark \\
\midrule
\multicolumn{12}{l}{\textit{Image--text retrieval datasets}} \\
\quad MS-COCO~{\scriptsize\citep{lin2014coco}}
    & 123K   & \xmark & \cmark & \cmark & \xmark & \xmark & \xmark & \xmark & \xmark & \xmark & \xmark \\
\quad Flickr30K~{\scriptsize\citep{young2014flickr30k}}
    & 31K    & \xmark & \cmark & \xmark & \cmark & \xmark & \xmark & \xmark & \xmark & \xmark & \xmark \\
\quad LAION-5B~{\scriptsize\citep{schuhmann2022laion}}
    & 5.85B  & \xmark & \cmark & \xmark & \cmark & \cmark & \xmark & \xmark & \xmark & \pmark & \xmark \\
\midrule
\multicolumn{12}{l}{\textit{Urban perception datasets}} \\
\quad Place Pulse 2.0~{\scriptsize\citep{dubey2016deeplearning}}
    & 111K   & \pmark & \xmark & \xmark & \xmark & \cmark & \xmark & \xmark & \xmark & \xmark & \xmark \\
\quad MMS-VPR~{\scriptsize\citep{ou2025mmsvpr}}
    & 110.5K & \cmark & \pmark & \xmark & \cmark & \xmark & \cmark & \xmark & \cmark & \cmark & \xmark \\
\quad UrbanFeel~{\scriptsize\citep{he2025urbanfeel}}
    & 14.3K  & \cmark & \xmark & \xmark & \pmark & \cmark & \cmark & \xmark & \xmark & \pmark & \xmark \\
\midrule
\rowcolor{oursrow}
\textbf{\ourname{} (Ours)}
    & \textbf{2M}
    & \cmark & \cmark & \cmark
    & \cmark & \cmark & \cmark
    & \cmark & \cmark & \cmark & \cmark \\
\bottomrule
\end{tabular}
}
\end{table*}

\section{Dataset Description}
\label{sec:dataset}

\subsection{Data Collection}
\label{sec:collection}

\begin{figure*}[t]
    \centering
    \includegraphics[width=\textwidth]{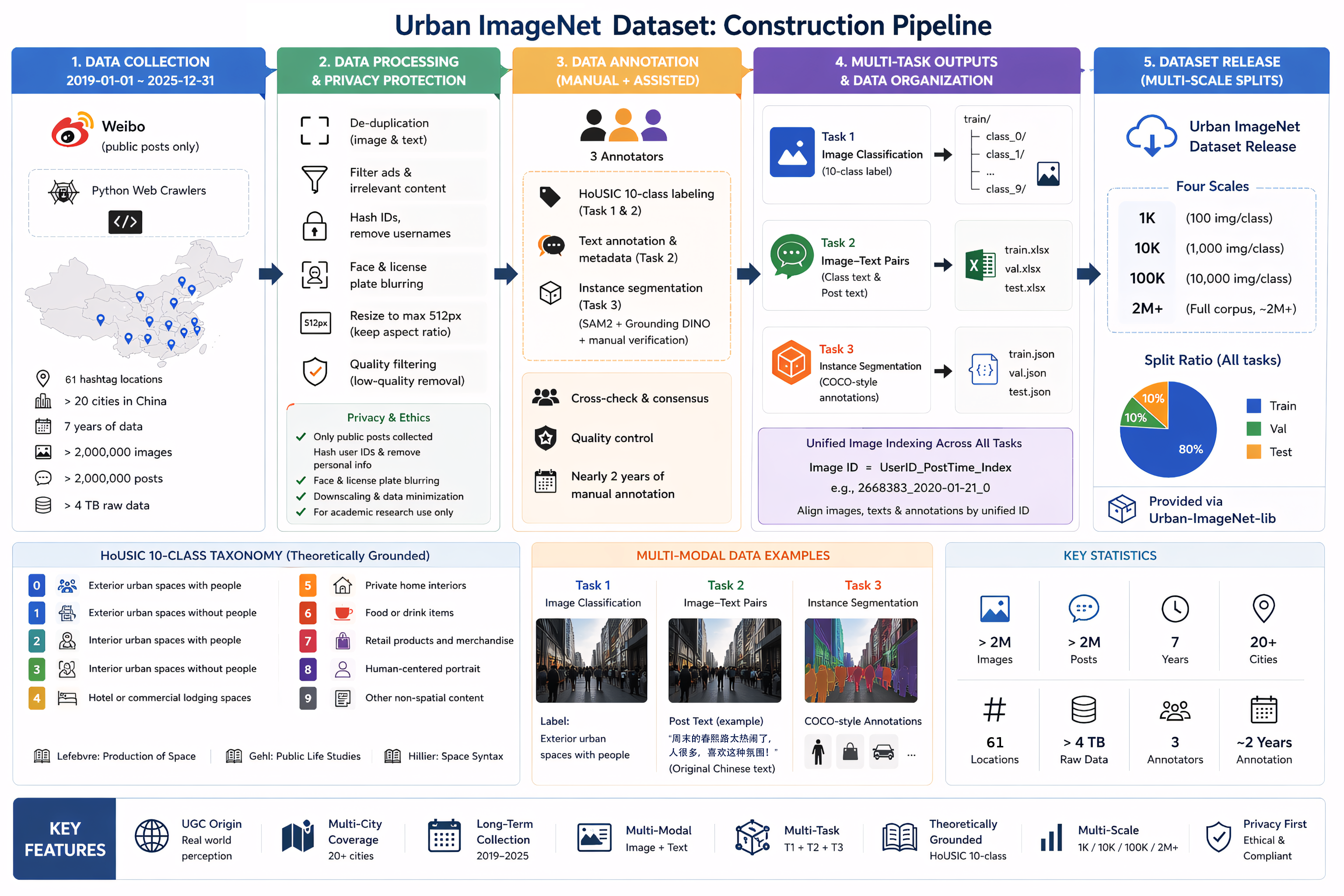}
    \caption{%
        Overview of the \ourname{} dataset construction and annotation pipeline.
    }
    \label{fig:pipeline}
\end{figure*}

As illustrated in Figure~\ref{fig:pipeline}, \ourname{} was constructed by
systematically crawling Sina Weibo using a Python-based web crawler targeting
location-specific hashtags at major urban commercial centres across China.
The crawler retrieved all publicly accessible posts published between
1~January~2019 and 31~December~2025, capturing image attachments (up to nine
per post), user-authored text, and post-level metadata, yielding a raw corpus
of more than \textbf{4~TB} and over \textbf{2~million} image--text pairs
across 61 hashtags spanning 24 cities in eight macro-regions.
Hashtag-driven crawling was preferred over GPS- or keyword-based alternatives
as Weibo hashtags are tightly coupled to identifiable physical locations,
yielding a site-specific corpus with minimal off-topic
contamination~\citep{hochman2013,boy2017}.
Site selection balanced geographic and typological diversity across China's
major regions, encompassing both enclosed shopping malls and open-block
commercial precincts; the complete site list and geographic distribution are
provided in Appendix~\ref{app:geo} (Table~\ref{tab:sites} and
Figure~\ref{fig:geo}).

\subsection{Data Processing and Privacy Protection}
\label{sec:processing}

\textbf{Data Cleaning and Curation.}
The raw corpus underwent a four-stage pipeline before annotation:
(1)~near-duplicate images were removed using perceptual hashing (pHash,
Hamming distance $\le 8$);
(2)~images smaller than $256{\times}256$ pixels were discarded;
(3)~a pre-trained NSFW classifier filtered inappropriate content; and
(4)~systematically repeated commercial advertisement posts were removed via
post-text hash similarity.

\textbf{Privacy Protection and Anonymisation.}
All original Weibo usernames were stripped and replaced with opaque numerical
identifiers; no username, profile picture, or account URL appears in the
released dataset.
Following the practice of large-scale street-level
datasets~\citep{anguelov2010streetview}, automated face detection,
licence-plate recognition, and QR-code detection were applied to all images,
with all detected regions blurred; a manual spot-check verified blurring
completeness.
The publicly released dataset contains images resized to a maximum side length
of 512~px, substantially reducing re-identification risk: the 100K subset
spans 6.15~GB at 512~px versus approximately 600~GB at original resolution.
The raw \textbf{4~TB} corpus is retained securely by the authors and will not
be publicly distributed. A full discussion of ethical considerations governing data collection, use, and distribution is provided in
Appendix~\ref{app:ethics}.

\subsection{The HUSIC 10-Class Annotation Framework}
\label{sec:husic}

\textbf{Motivation.}
Raw location-tagged social media content is inherently heterogeneous.
A user posting under a single urban-district hashtag ``\#Beijing Sanlitun'' produces a corpus
spanning architectural photography, dining imagery, merchandise display, selfies,
hotel promotion, and noise.
Without a principled classification framework, downstream spatial analyses
are confounded by this heterogeneity~\citep{hochman2013}. The \textbf{\husic{}} (\textbf{H}ierarchical \textbf{U}rban \textbf{S}pace
\textbf{I}mage \textbf{C}lassification) framework addresses this by
providing a complete, theoretically grounded taxonomy that (1)~identifies
and filters the specific image types relevant to each research question,
and (2)~provides a 10-way classification benchmark capturing the full
semantic spectrum of urban space social media content.

\textbf{Theoretical grounding.}
\husic{} is the first large-scale vision taxonomy whose class boundaries are
defined by domain-expert concepts rather than data-driven frequency, drawing
on three complementary bodies of urban theory.
\textit{(i) Lefebvre's Production of Space}~\citep{lefebvre1991}: Lefebvre's distinction between \emph{conceived space} (design intent) and
\emph{lived space} (social appropriation through use)~\citep{lefebvre1991}
motivates the \textit{with}/\textit{without people} axis within each spatial
group---a distinction absent from all existing vision benchmarks.  \textit{(ii) Gehl's Public Life Studies}~\citep{gehl2011}: Gehl's finding that social activity is both an indicator and a
self-reinforcing generator of successful public space~\citep{gehl2011}
justifies treating activated and non-activated spaces as analytically
distinct categories. \textit{(iii) Newman's Spatial Hierarchy and Publicity Gradient}~\citep{newman1972}:
Newman's defensible-space framework~\citep{newman1972}, which
conceptualises urban environments along a public-to-private gradient,
provides the conceptual basis for \husic{}'s three-tier spatial hierarchy:
publicly accessible spaces (e.g., exterior plazas, commercial interiors),
transitional semi-public spaces (e.g., hotel lobbies), and privately
controlled spaces (e.g., residential interiors).

\textbf{Dual function.}
\husic{} serves as both a \emph{UGC filtering pipeline} (extract relevant
image subsets from raw social media) and a \emph{classification benchmark}
(10-way T1 task measuring full-spectrum urban content understanding).
Full class definitions in Table~\ref{tab:husic} (Appendix~\ref{app:husic}).

\subsection{Annotation and Dataset Organisation}
\label{sec:annotation}

\textbf{Manual annotation (T1 and T2).}
The 100K balanced benchmark set was manually annotated by three trained researchers following a standardized guideline.
A shared 3,000-image subset was double-annotated for agreement, yielding Cohen's $\kappa=0.87$, commonly interpreted as almost perfect agreement \citep{landis1977measurement}.
Disagreements were resolved by majority vote and guideline revision.

\textbf{T1 file structure.}
Each split (\texttt{train/val/test}) contains 10 class-named subdirectories;
images in each subdirectory carry that class as ground truth.
Integer labels 0--9 follow lexicographic sort, compatible with PyTorch
\texttt{ImageFolder}.
Split ratio: 80:10:10 across all four tiers.

\textbf{T2 multi-modal structure.}
Each split is accompanied by a metadata spreadsheet
(\texttt{train.xlsx}, \texttt{val.xlsx}, \texttt{test.xlsx})
with 15 columns per record (full schema in Appendix~\ref{app:t2schema}).
The \texttt{Image~Filename} field (\texttt{UserID\_PostTime\_Index}) is
the primary join key linking images to post text.
One post may contain up to nine images; all images from the same post share
the same \texttt{Post~ID} and \texttt{Post~Text}.
This one-to-many structure (one text $\to$ multiple images) is intrinsic
to the Weibo platform and defines the multi-positive retrieval protocol
described in Section~\ref{sec:task2}.

\textbf{T3 instance segmentation.}
Annotations were generated via Grounding~DINO~\citep{liu2023groundingdino}
+ SAM~2~\citep{ravi2024sam2} with \emph{per-class text prompts}
(16--20 semantically appropriate object terms per \husic{} class;
full vocabulary in Appendix~\ref{app:segvocab}).
Training pseudo-labels use confidence $\ge 0.35$ and IoU $\ge 0.80$.
The evaluation subset applies stricter thresholds
($\ge 0.50$, $\ge 0.88$) with human review, ensuring reliable
ground truth for model comparison.
Annotations stored as COCO-compatible JSON.
T3 evaluation adopts a \emph{class-agnostic} protocol, treating all
detected objects as a single \texttt{object} category to obtain
conservative, architecture-comparable metrics uncorrupted by
class-imbalanced pseudo-labels.

\textbf{Multi-scale tiers.}
Four tiers are released: \textbf{1K} (100 images/class),
\textbf{10K} (1,000/class), \textbf{100K} (10,000/class; strictly balanced;
$\approx$6.15\,GB), and \textbf{2M} (full corpus; class-imbalanced).
All three tasks share identical image files across tiers; labels and splits
are consistent.

\section{Benchmark Tasks and Evaluation Protocol}
\label{sec:benchmark}

\begin{figure}[t]
  \centering
  \includegraphics[width=\linewidth]{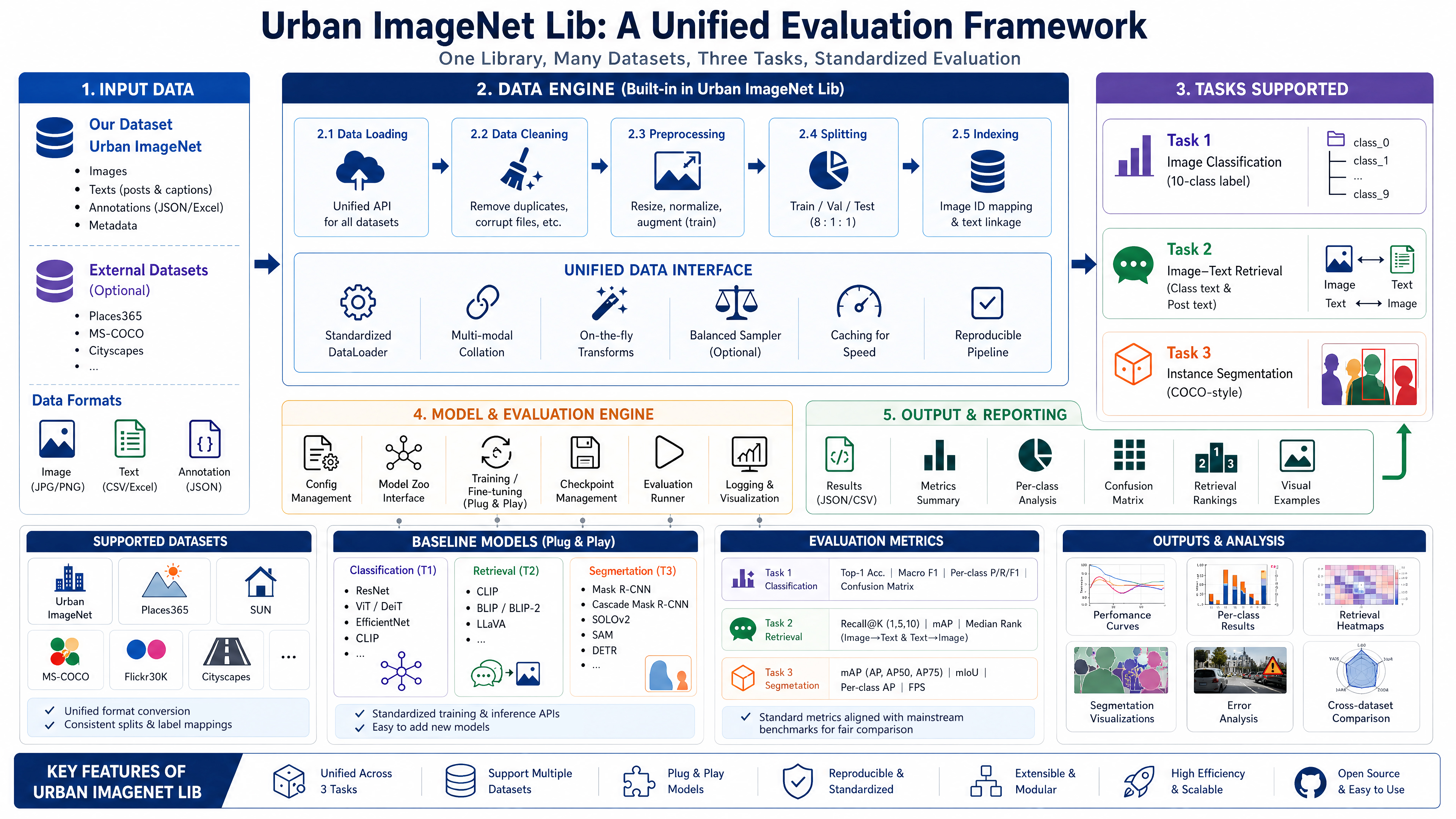}
  \caption{\ourname{}-lib architecture, supporting T1, T2, and T3 in a single
  unified framework.}
  \label{fig:netlib}
\end{figure}

As illustrated in Figure~\ref{fig:netlib}, \ourname{} is accompanied by
\textbf{\ourname{}-lib}, a reproducible benchmarking library providing
modular data loaders, fine-tuning pipelines, evaluation scripts, and
cross-dataset adapters for direct comparison with Places365, MS-COCO, and
Cityscapes.

\subsection{Task 1: Urban Scene Semantic Classification}
\label{sec:task1}
\textbf{Setup:} Given an image, predict its \husic{} label (0--9).
Fine-tuned on 80K training images; evaluated on 10K test split with
five-fold cross-validation.
\textbf{Baselines:}
ResNet-\{18/50/152\}~\citep{he2016resnet},
EfficientNet-B4~\citep{tan2019efficientnet},
ViT-B/16~\citep{dosovitskiy2021vit},
DeiT-B~\citep{touvron2021deit},
CLIP ViT-L/14 (zero-shot + fine-tuned)~\citep{radford2021clip}.
\textbf{Metrics:} Top-1 Accuracy, Macro-F1, per-class P/R/F1.

\subsection{Task 2: Cross-Modal Image--Text Retrieval}
\label{sec:task2}

Task~2 evaluates two sub-configurations reflecting the dataset's two
textual modalities:

\textbf{T2-1 (Category-Level Retrieval).}

Text queries are the ten \husic{} class names, formatted as
``\texttt{This is a photo of \{class\_name\}}'' for CLIP-style models.
For LLaVA-style models, visual question prompts ask which \husic{} class best
describes the image.
This sub-task measures zero-shot \emph{urban semantic alignment}: whether a
model's joint embedding space captures the theoretical distinctions encoded in
\husic{}.
All images from a queried class are treated as correct
(\emph{many-relevant} retrieval).

\textbf{T2-2 (Post-Level Retrieval).}
Queries are original Weibo post texts (Chinese; $\approx$50--500 characters).
Because one Weibo post may include up to nine images, the positive set for
a given text query is \emph{all images from that post}
($\approx$1.06 images per post on average over the 10K test split,
yielding $\sim$943 unique text queries over 1,000 test images).
Retrieval is evaluated in both directions:
\textit{T2I} (text $\to$ image): given a post text, rank all test images,
report R@$K$ if any positive image appears in top-$K$;
\textit{I2T} (image $\to$ text): given an image, rank all unique post texts,
report R@$K$ if the correct post appears in top-$K$.
Both directions employ a \emph{multi-positive} protocol: a query is deemed
a hit at rank $K$ when any one of its positive targets appears within
top-$K$ retrieved results.
Random-chance baselines: T2I R@1\,$\approx$\,0.106\%
(1.06 positives / 1K test images);
I2T R@1\,$\approx$\,0.106\% (1 positive / 943 unique texts).
\textbf{Baselines:} CLIP (zero-shot + fine-tuned)~\citep{radford2021clip},
BLIP~\citep{li2022blip}, BLIP-2~\citep{li2023blip2}.
\textbf{Metrics:} Recall@$K$ ($K\in\{1,5,10\}$), mAP, Median Rank.
Full evaluation code provided in Appendix~\ref{app:t2code}.

\subsection{Task 3: Instance Segmentation}
\label{sec:task3}

\textbf{Setup.}
Evaluate on the human-verified evaluation subset (10K images) under
a \emph{class-agnostic} protocol (all object instances merged into a single
\texttt{object} category) to obtain robust, architecture-comparable metrics.
\textbf{Baselines:} Mask~R-CNN~\citep{he2017maskrcnn},
Cascade~Mask~R-CNN~\citep{cai2019cascade}
(both trained on 10K pseudo-label annotations);
SAM box-refinement variants (Cascade+SAM and Mask~R-CNN+SAM), in which
predicted bounding boxes from the respective detectors are used as prompts
for zero-shot SAM inference, bypassing SAM fine-tuning entirely;
SAM zero-shot with GT box prompt (oracle upper bound)~\citep{kirillov2023sam}.
\textbf{Metrics:} Mask AP, AP$_{50}$, AP$_{75}$, mIoU, FPS.

\section{Experimental Results}
\label{sec:experiments}

\subsection{Task 1 Results: Urban Scene Semantic Classification}
\label{sec:exp_t1}

Table~\ref{tab:t1_results} presents results on the 100K benchmark
(80K/10K/10K split, five-fold cross-validation). EfficientNet-B4 achieves the best classification performance at 84.9\% Top-1 accuracy and 84.9\% Macro-F1.
The CNN and transformer classifiers are close on the balanced 10-way task, with ResNet-152 and DeiT-B reaching roughly 80\% Top-1 accuracy.
CLIP zero-shot performs poorly because \husic{} labels such as activated exterior space or non-activated interior space are not simple web categories.
Fine-tuning improves CLIP substantially, but it remains below the supervised visual classifiers in the primary setup.
Per-class results in Appendix~\ref{app:full_results} show that interior-without-people and exterior/interior activation boundaries are the most difficult categories.

\begin{table}[t]
\centering
\caption{Task 1 urban scene semantic classification results on the balanced benchmark split. 
}
\label{tab:t1_results}
\setlength{\tabcolsep}{6pt}
\begin{tabular}{lcc}
\toprule
\textbf{Model} & \textbf{Top-1 Acc.} & \textbf{Macro-F1} \\
\midrule
ResNet-18 & 75.9 & 75.4 \\
ResNet-50 & 79.7 & 79.9 \\
ResNet-152 & 80.5 & 80.4 \\
EfficientNet-B4 & \textbf{84.9} & \textbf{84.9} \\
ViT-B/16 & 79.0 & 79.0 \\
DeiT-B & 80.3 & 80.2 \\
CLIP (zero-shot) & 37.9 & 35.0 \\
CLIP (fine-tuned) & 69.1 & 67.5 \\
\bottomrule
\end{tabular}
\end{table}

\subsection{Task~2: Cross-Modal Retrieval}
\label{sec:t2_results}

\begin{table*}[t]
\centering
\caption{Task~2 retrieval results.
ZS\,=\,zero-shot; FT\,=\,fine-tuned. MedR: lower is better.}
\label{tab:t2_results}
\setlength{\tabcolsep}{4.5pt}
\renewcommand{\arraystretch}{1.08}
\small
\begin{tabular}{llccccc}
\toprule
\textbf{Setting} & \textbf{Model} & \textbf{R@1} & \textbf{R@5} &
\textbf{R@10} & \textbf{mAP} & \textbf{MedR} \\
\midrule
\multirow{4}{*}{Category label}
  & CLIP ZS  & 54.2 & 96.5 & 100.0 & 53.3 & 1.5 \\
  & CLIP FT  & 92.7 & 99.8 & 100.0 & 90.7 & 1.0 \\
  & BLIP ZS  & 14.9 & 43.6 &  80.0 & 19.8 & 6.2 \\
  & BLIP FT  & \textbf{94.2} & \textbf{99.8} & \textbf{100.0} & \textbf{93.3} & \textbf{1.0} \\
\midrule
\multirow{4}{*}{Post text}
  & CLIP ZS  &  2.6 &  5.4 &  7.0 &  4.5 & 328 \\
  & CLIP FT  & \textbf{8.1} & \textbf{16.9} & \textbf{23.5} & \textbf{13.2} & \textbf{64} \\
  & BLIP ZS  &  0.1 &  0.4 &  1.2 &  0.8 & 477 \\
  & BLIP FT  &  1.9 &  6.8 & 11.6 &  5.5 &  92 \\
\midrule
\multirow{4}{*}{Post\,+\,label}
  & CLIP ZS  &  2.7 &  6.2 &  9.4 &  5.5 & 128 \\
  & CLIP FT  & \textbf{9.3} & \textbf{22.8} & \textbf{32.3} & \textbf{17.0} & \textbf{25} \\
  & BLIP ZS  &  0.1 &  0.6 &  1.2 &  0.8 & 469 \\
  & BLIP FT  &  2.6 &  9.2 & 16.7 &  8.0 &  38 \\
\bottomrule
\end{tabular}
\end{table*}

\begin{figure}[t]
  \centering
  \includegraphics[width=0.8\linewidth]{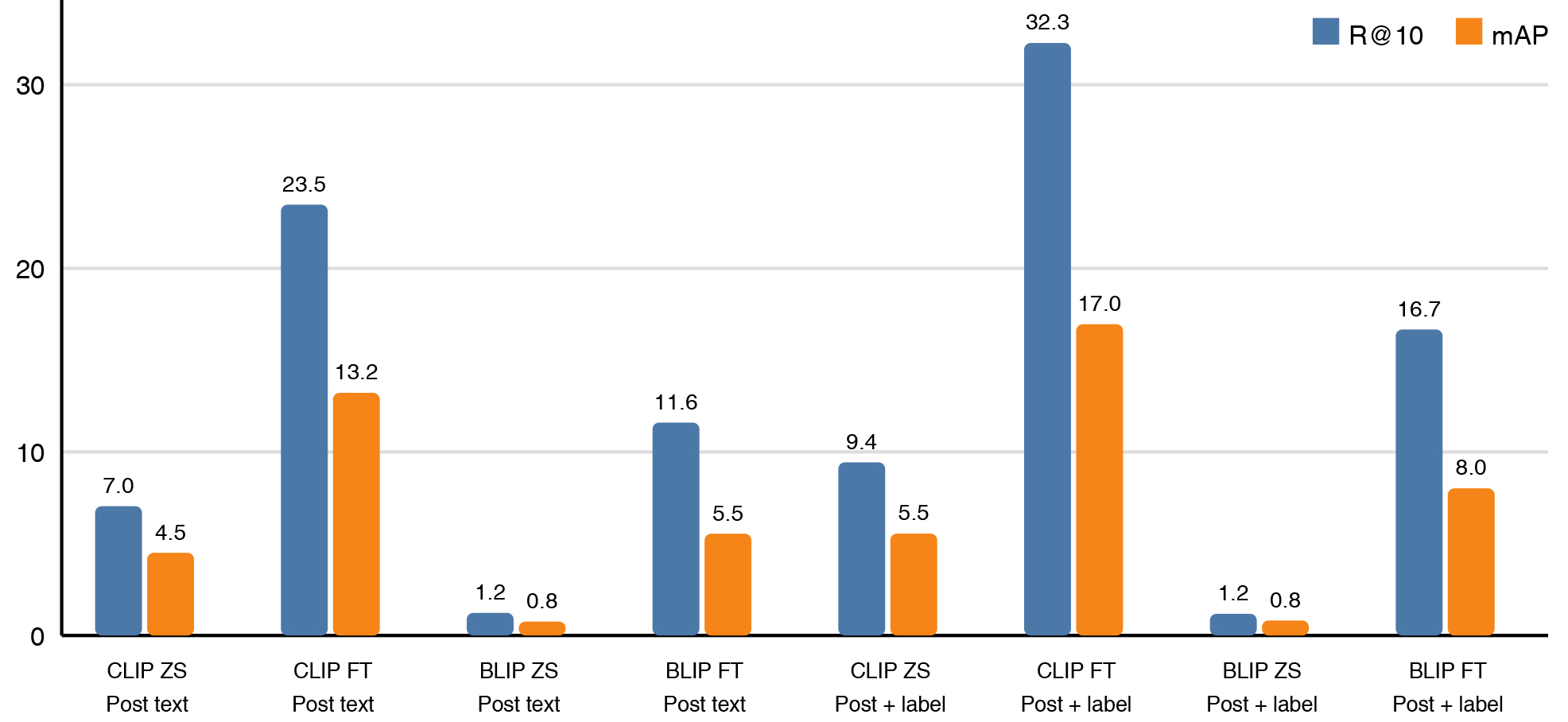}
  \caption{Task~2 retrieval results (avg.\ T2I\,+\,I2T).
  Category-label retrieval (left) is near-trivial after fine-tuning
  ($\ge$92\% R@1); post-text retrieval (right) remains genuinely hard
  ($\le$8\% R@1).}
  \label{fig:t2_results}
\end{figure}

Results are shown in Table~\ref{tab:t2_results} and Figure~\ref{fig:t2_results}.
Category-label retrieval is straightforward: fine-tuned BLIP and CLIP reach
94.2\% and 92.7\% average R@1, confirming that \husic{} descriptions provide
strong cross-modal signal; zero-shot CLIP already achieves 54.2\%.
Post-text retrieval is substantially harder, as Weibo posts are short informal
narratives (median 32 characters) rather than image descriptions, and a single
post may accompany images spanning multiple \husic{} classes.
Against a random-chance baseline of $\approx$0.1\% R@1, fine-tuned CLIP
achieves 8.1\% (${\approx}76\times$ chance), the best result among all models;
appending the \husic{} label (Post\,+\,label) raises this further to 9.3\%
R@1 and 32.3\% R@10, serving as a metadata-assisted upper bound.
The low post-level scores reflect an intrinsic property of
social media text, establishing a concrete target for future urban-domain
vision--language models.

\subsection{Task~3: Instance Segmentation}
\label{sec:exp_t3}

Table~\ref{tab:t3_main} presents T3 results.
Cascade Mask R-CNN achieves the best standalone AP (0.290), outperforming
Mask R-CNN (0.267).
SAM box-refinement substantially improves both: Mask R-CNN+SAM reaches
AP\,=\,0.373 ($+40\%$ relative) and Cascade+SAM reaches 0.369 ($+27\%$),
with the weaker base detector producing the marginally better refined result,
suggesting that Mask R-CNN boxes provide richer spatial diversity as SAM
prompts.
The GT-box SAM oracle (AP\,=\,0.749) establishes a clear upper bound for
future domain-specific detection progress; as pseudo-labels were generated
by SAM, this score partly reflects circularity, and the SAM-refinement
results trained on noisy pseudo-labels provide the more informative measure
of generalizable performance.

\begin{table}[t]
\centering
\caption{\textbf{T3 instance segmentation results} 
  $^\dagger$GT-box SAM uses ground-truth bounding boxes---an oracle
  upper bound, not a trainable baseline.
  \textbf{Bold}\,=\,best trainable model.}
\label{tab:t3_main}
\setlength{\tabcolsep}{4pt}
\renewcommand{\arraystretch}{1.08}
\small
\begin{tabular}{l c c c c c}
\toprule
\textbf{Model} & \textbf{AP} & \textbf{AP}$_{50}$ & \textbf{AP}$_{75}$ &
\textbf{mIoU} & \textbf{FPS} \\
\midrule
Mask R-CNN~\citep{he2017maskrcnn}           & 0.267 & 0.472 & 0.276 & 0.629 & 15.4 \\
Cascade Mask R-CNN~\citep{cai2019cascade}   & 0.290 & 0.495 & 0.299 & 0.635 & 12.7 \\
\midrule
Mask R-CNN + SAM~\citep{kirillov2023sam}    & \textbf{0.373} & 0.563 & \textbf{0.378} & ---  & $\approx$0 \\
\textbf{Cascade + SAM}~\citep{kirillov2023sam,cai2019cascade}
                                            & 0.369 & \textbf{0.531} & 0.380 & ---  & $\approx$0 \\
\midrule
GT-box SAM (oracle)$^\dagger$~\citep{kirillov2023sam} &
  0.749 & 0.924 & 0.805 & --- & --- \\
\bottomrule
\end{tabular}
\end{table}

\begin{figure}[t]
  \centering
  \includegraphics[width=0.9\linewidth]{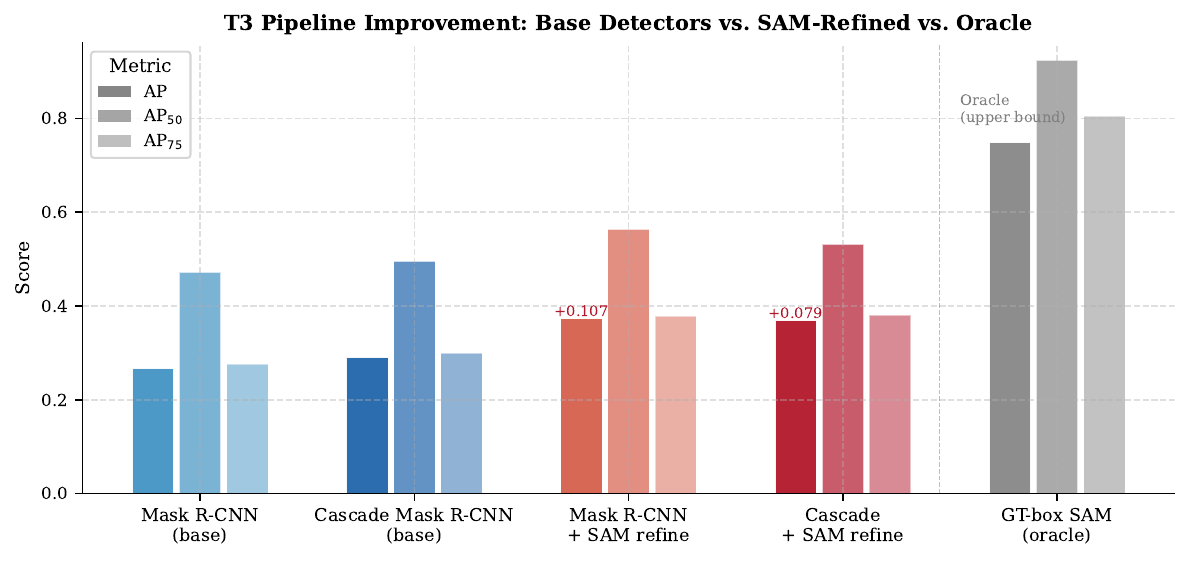}
  \caption{\textbf{T3 pipeline improvement.}
    Adding SAM box-refinement to both detectors substantially increases AP.
    Mask R-CNN+SAM achieves the highest trainable AP (0.373), providing
    a strong open-source baseline for future work.
    Lighter bars: AP$_{50}$; lightest: AP$_{75}$.}
  \label{fig:t3_pipeline}
\end{figure}

\begin{figure*}[t]
  \centering
  \includegraphics[width=\linewidth]{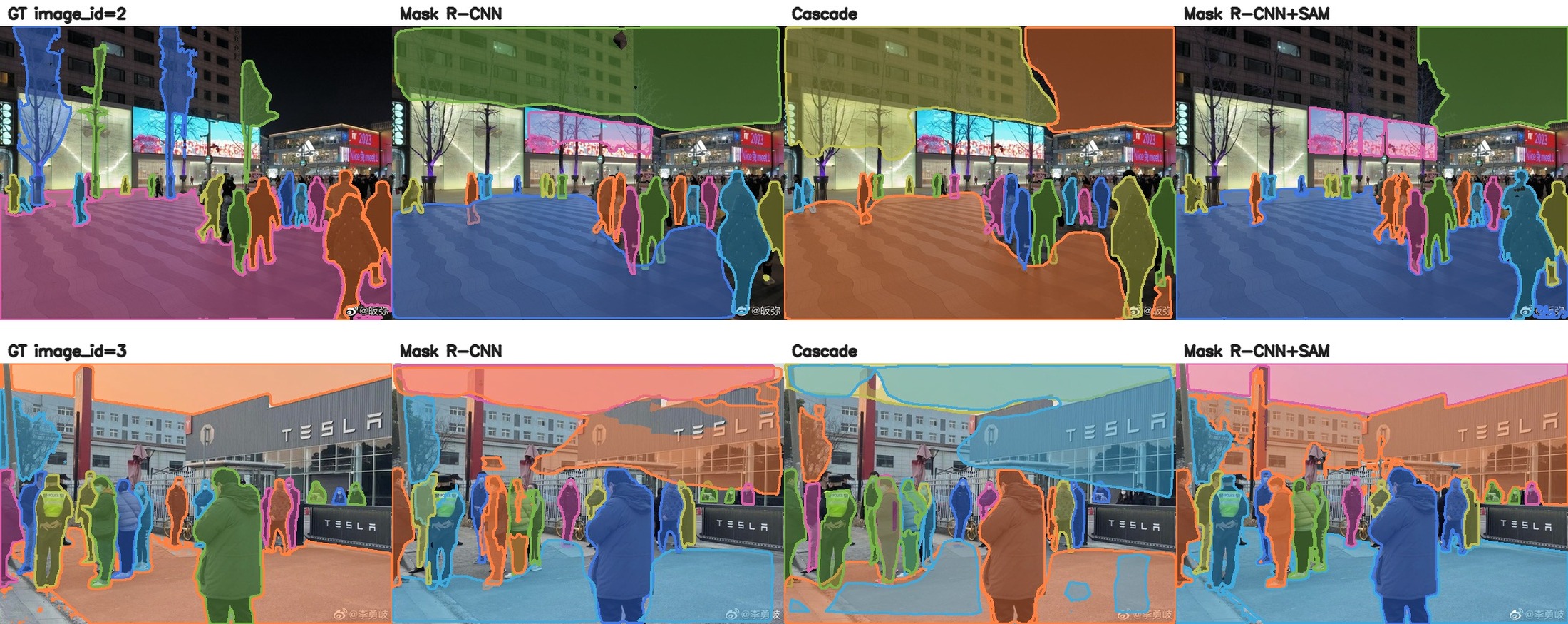}
  \caption{\textbf{Task 3 qualitative segmentation examples.}  Colour-coded instance masks from Mask R-CNN, Cascade Mask R-CNN, and Mask R-CNN+SAM. Left: ground truth; right: predictions.
  }
  \label{fig:seg_examples}
\end{figure*}

\section{Discussion}
\label{sec:discussion}

\textbf{Task-dependent architecture selection.}
No single model dominates all three tasks: EfficientNet-B4 leads T1;
CLIP fine-tuned leads T2-Post; Cascade Mask R-CNN with SAM refinement leads
T3---confirming that urban multi-task benchmarking requires task-appropriate
architecture selection.

\textbf{Domain gap.}
All zero-shot baselines perform substantially below their fine-tuned
counterparts, most severely for T2-Post, where colloquial Chinese  post texts differ fundamentally from English captions.

\textbf{SAM refinement.}
The SAM box-refinement pipeline requires no additional training yet yields
a 27--40\% relative AP improvement over its base detector, advocating for
foundation-model post-processing as a practical, training-free strategy for
domain-adapted detection pipelines.

\textbf{Scaling behaviour.}
All models improve monotonically across the 1K, 10K, and 100K tiers, with the
1K$\to$10K gain (10--12\%), and 10K$\to$100K (5\%),
consistent with standard scaling laws~\citep{kaplan2020scaling}.
Detailed scaling behaviour results and discussion of
limitations are provided in Appendix~\ref{app:scaling} and
Appendix~\ref{app:discussion}.

\section{Conclusion}
\label{sec:conclusion}

We presented \ourname{}, a large-scale multi-modal dataset of over
\textbf{2M} user-generated image--text pairs from Weibo, covering 24 Chinese
cities, 61 venues, and seven years (2019--2025).
The \husic{} 10-class framework, grounded in Lefebvre, Gehl, and Newman,
provides theoretically principled annotations supporting T1 scene
classification, T2 cross-modal retrieval, and T3 instance segmentation within
a single unified benchmark.
Experiments show that domain fine-tuning consistently improves all baselines,
that post-level retrieval on social media text is a genuinely hard open
problem, and that SAM-based refinement offers a strong training-free gain for
instance segmentation.
Beyond the dataset itself, \ourname{} is designed to support clearly scoped
evaluative claims: each benchmark split, annotation protocol, and evaluation
metric is documented with its underlying assumptions and interpretive
boundaries, consistent with rigorous evaluation science.
\ourname{} thereby provides a reproducible, theoretically grounded foundation
for social media-based urban space perception research, and a concrete example
of how domain-specific datasets can be constructed to advance both model
capabilities and evaluation methodology.

\clearpage

\bibliographystyle{plainnat}
\bibliography{urban_imagenet_refs}

@book{gehl2011,
	title        = {Life Between Buildings: Using Public Space},
	author       = {Gehl, Jan},
	year         = 2011,
	publisher    = {Island Press},
	address      = {Washington, DC},
	edition      = {6th}
}

@book{whyte1980,
	title        = {The Social Life of Small Urban Spaces},
	author       = {Whyte, William H.},
	year         = 1980,
	publisher    = {Conservation Foundation},
	address      = {Washington, DC}
}

@book{lefebvre1991,
	title        = {The Production of Space},
	author       = {Lefebvre, Henri},
	year         = 1991,
	publisher    = {Blackwell},
	address      = {Oxford},
	note         = {Translated by D. Nicholson-Smith}
}

@book{hillier1984,
	title        = {The Social Logic of Space},
	author       = {Hillier, Bill and Hanson, Julienne},
	year         = 1984,
	publisher    = {Cambridge University Press},
	address      = {Cambridge}
}

@book{lynch1960,
	title        = {The Image of the City},
	author       = {Lynch, Kevin},
	year         = 1960,
	publisher    = {MIT Press},
	address      = {Cambridge, MA}
}

@article{hochman2013,
	title        = {Zooming into an {Instagram} City: Reading the Local through Social Media},
	author       = {Hochman, Nadav and Manovich, Lev},
	year         = 2013,
	journal      = {First Monday},
	volume       = 18,
	number       = 7,
	doi          = {10.5210/fm.v18i7.4711}
}

@article{boy2017,
	title        = {Reassembling the City through {Instagram}},
	author       = {Boy, John D. and Uitermark, Justus},
	year         = 2017,
	journal      = {Transactions of the Institute of British Geographers},
	volume       = 42,
	number       = 2,
	pages        = {612--624},
	doi          = {10.1111/tran.12185}
}

@inproceedings{dubey2016deeplearning,
	title        = {Deep Learning the City: Quantifying Urban Perception at a Global Scale},
	author       = {Dubey, Abhimanyu and Naik, Nikhil and Parikh, Devi and Raskar, Ramesh and Hidalgo, C{\'e}sar A.},
	year         = 2016,
	booktitle    = {European Conference on Computer Vision (ECCV)},
	publisher    = {Springer},
	pages        = {196--212},
	doi          = {10.1007/978-3-319-46448-0_12}
}

@article{zhou2018places,
	title        = {Places: A 10 Million Image Database for Scene Recognition},
	author       = {Zhou, Bolei and Lapedriza, Agata and Khosla, Aditya and Oliva, Aude and Torralba, Antonio},
	year         = 2018,
	journal      = {IEEE Transactions on Pattern Analysis and Machine Intelligence},
	volume       = 40,
	number       = 6,
	pages        = {1452--1464},
	doi          = {10.1109/TPAMI.2017.2723009}
}

@inproceedings{xiao2010sun,
	title        = {{SUN} Database: Large-Scale Scene Recognition from Abbey to Zoo},
	author       = {Xiao, Jianxiong and Hays, James and Ehinger, Krista A. and Oliva, Aude and Torralba, Antonio},
	year         = 2010,
	booktitle    = {IEEE Conference on Computer Vision and Pattern Recognition (CVPR)},
	pages        = {3485--3492},
	doi          = {10.1109/CVPR.2010.5539970}
}

@inproceedings{quattoni2009indoor,
	title        = {Recognizing Indoor Scenes},
	author       = {Quattoni, Ariadna and Torralba, Antonio},
	year         = 2009,
	booktitle    = {IEEE Conference on Computer Vision and Pattern Recognition (CVPR)},
	pages        = {413--420},
	doi          = {10.1109/CVPR.2009.5206537}
}

@inproceedings{cordts2016cityscapes,
	title        = {The Cityscapes Dataset for Semantic Urban Scene Understanding},
	author       = {Cordts, Marius and Omran, Mohamed and Ramos, Sebastian and Rehfeld, Timo and Enzweiler, Markus and Benenson, Rodrigo and Franke, Uwe and Roth, Stefan and Schiele, Bernt},
	year         = 2016,
	booktitle    = {IEEE Conference on Computer Vision and Pattern Recognition (CVPR)},
	pages        = {3213--3223},
	doi          = {10.1109/CVPR.2016.350}
}

@article{zhou2019ade20k,
	title        = {Semantic Understanding of Scenes through the {ADE20K} Dataset},
	author       = {Zhou, Bolei and Zhao, Hang and Puig, Xavier and Xiao, Tete and Fidler, Sanja and Barriuso, Adela and Torralba, Antonio},
	year         = 2019,
	journal      = {International Journal of Computer Vision},
	volume       = 127,
	number       = 3,
	pages        = {302--321},
	doi          = {10.1007/s11263-018-1140-0}
}

@inproceedings{gupta2019lvis,
	title        = {{LVIS}: A Dataset for Large Vocabulary Instance Segmentation},
	author       = {Gupta, Agrim and Dollar, Piotr and Girshick, Ross},
	year         = 2019,
	booktitle    = {IEEE Conference on Computer Vision and Pattern Recognition (CVPR)},
	pages        = {5356--5364},
	doi          = {10.1109/CVPR.2019.00550}
}

@inproceedings{neuhold2017mapillary,
	title        = {The Mapillary Vistas Dataset for Semantic Understanding of Street Scenes},
	author       = {Neuhold, Gerhard and Ollmann, Tobias and Bul{\`o}, Samuel Rota and Kontschieder, Peter},
	year         = 2017,
	booktitle    = {IEEE International Conference on Computer Vision (ICCV)},
	pages        = {4990--4999},
	doi          = {10.1109/ICCV.2017.534}
}

@inproceedings{lin2014coco,
	title        = {Microsoft {COCO}: Common Objects in Context},
	author       = {Lin, Tsung-Yi and Maire, Michael and Belongie, Serge and Hays, James and Perona, Pietro and Ramanan, Deva and Doll{\'a}r, Piotr and Zitnick, C. Lawrence},
	year         = 2014,
	booktitle    = {European Conference on Computer Vision (ECCV)},
	publisher    = {Springer},
	pages        = {740--755},
	doi          = {10.1007/978-3-319-10602-1_48}
}

@article{young2014flickr30k,
	title        = {From Image Descriptions to Visual Denotations: New Similarity Metrics for Semantic Inference over Event Descriptions},
	author       = {Young, Peter and Lai, Alice and Hodosh, Micah and Hockenmaier, Julia},
	year         = 2014,
	journal      = {Transactions of the Association for Computational Linguistics},
	volume       = 2,
	pages        = {67--78},
	doi          = {10.1162/tacl_a_00166}
}

@inproceedings{schuhmann2022laion,
	title        = {{LAION-5B}: An Open Large-Scale Dataset for Training Next Generation Image-Text Models},
	author       = {Schuhmann, Christoph and Beaumont, Romain and Vencu, Richard and Gordon, Cade and Wightman, Ross and Cherti, Mehdi and Coombes, Theo and Katta, Aarush and Mullis, Clayton and Wortsman, Mitchell and Schramowski, Patrick and Kundurthy, Srivatsa and Crowson, Katherine and Schmidt, Ludwig and Beaumont, Romain and Jitsev, Jenia},
	year         = 2022,
	booktitle    = {Advances in Neural Information Processing Systems (NeurIPS)},
	volume       = 35,
	pages        = {25278--25294}
}

@inproceedings{radford2021clip,
	title        = {Learning Transferable Visual Models from Natural Language Supervision},
	author       = {Radford, Alec and Kim, Jong Wook and Hallacy, Chris and Ramesh, Aditya and Goh, Gabriel and Agarwal, Sandhini and Sastry, Girish and Askell, Amanda and Mishkin, Pamela and Clark, Jack and Krueger, Gretchen and Sutskever, Ilya},
	year         = 2021,
	booktitle    = {International Conference on Machine Learning (ICML)},
	pages        = {8748--8763}
}

@inproceedings{li2022blip,
	title        = {{BLIP}: Bootstrapping Language-Image Pre-training for Unified Vision-Language Understanding and Generation},
	author       = {Li, Junnan and Li, Dongxu and Xiong, Caiming and Hoi, Steven},
	year         = 2022,
	booktitle    = {International Conference on Machine Learning (ICML)},
	pages        = {12888--12900}
}

@inproceedings{li2023blip2,
	title        = {{BLIP-2}: Bootstrapping Language-Image Pre-training with Frozen Image Encoders and Large Language Models},
	author       = {Li, Junnan and Li, Dongxu and Savarese, Silvio and Hoi, Steven},
	year         = 2023,
	booktitle    = {International Conference on Machine Learning (ICML)}
}

@inproceedings{liu2024llava,
	title        = {Visual Instruction Tuning ({LLaVA})},
	author       = {Liu, Haotian and Li, Chunyuan and Wu, Qingyang and Lee, Yong Jae},
	year         = 2024,
	booktitle    = {Advances in Neural Information Processing Systems (NeurIPS)}
}

@inproceedings{kirillov2023sam,
	title        = {Segment Anything},
	author       = {Kirillov, Alexander and Mintun, Eric and Ravi, Nikhila and Mao, Hanzi and Rolland, Chloe and Gustafson, Laura and Xiao, Tete and Whitehead, Spencer and Berg, Alexander C. and Lo, Wan-Yen and Doll{\'a}r, Piotr and Girshick, Ross},
	year         = 2023,
	booktitle    = {IEEE International Conference on Computer Vision (ICCV)},
	pages        = {4015--4026},
	doi          = {10.1109/ICCV51070.2023.00371}
}

@article{ravi2024sam2,
	title        = {{SAM~2}: Segment Anything in Images and Videos},
	author       = {Ravi, Nikhila and Gabeur, Valentin and Hu, Yuan-Ting and Hu, Ronghang and Ryali, Chaitanya and Ma, Tengyu and Khedr, Haitham and R{\"a}dle, Roman and Rolland, Chloe and Gustafson, Laura and Mintun, Eric and Pan, Junting and Alwala, Kalyan Vasudev and Carion, Nicolas and Wu, Chao-Yuan and Girshick, Ross and Doll{\'a}r, Piotr and Feichtenhofer, Christoph},
	year         = 2024,
	journal      = {arXiv preprint arXiv:2408.00714}
}

@article{liu2023groundingdino,
	title        = {Grounding {DINO}: Marrying {DINO} with Grounded Pre-Training for Open-Set Object Detection},
	author       = {Liu, Shilong and Zeng, Zhaoyang and Ren, Tianhe and Li, Feng and Zhang, Hao and Yang, Jie and Li, Chunyuan and Yang, Jianwei and Su, Hang and Zhu, Jun and Zhang, Lei},
	year         = 2023,
	journal      = {arXiv preprint arXiv:2303.05499}
}

@inproceedings{he2017maskrcnn,
	title        = {Mask {R-CNN}},
	author       = {He, Kaiming and Gkioxari, Georgia and Doll{\'a}r, Piotr and Girshick, Ross},
	year         = 2017,
	booktitle    = {IEEE International Conference on Computer Vision (ICCV)},
	pages        = {2961--2969},
	doi          = {10.1109/ICCV.2017.322}
}

@article{cai2019cascade,
	title        = {Cascade {R-CNN}: High Quality Object Detection and Instance Segmentation},
	author       = {Cai, Zhaowei and Vasconcelos, Nuno},
	year         = 2021,
	journal      = {IEEE Transactions on Pattern Analysis and Machine Intelligence},
	volume       = 43,
	number       = 5,
	pages        = {1483--1498},
	doi          = {10.1109/TPAMI.2019.2956516}
}

@inproceedings{he2016resnet,
	title        = {Deep Residual Learning for Image Recognition},
	author       = {He, Kaiming and Zhang, Xiangyu and Ren, Shaoqing and Sun, Jian},
	year         = 2016,
	booktitle    = {IEEE Conference on Computer Vision and Pattern Recognition (CVPR)},
	pages        = {770--778},
	doi          = {10.1109/CVPR.2016.90}
}

@inproceedings{dosovitskiy2021vit,
	title        = {An Image is Worth {16$\times$16} Words: Transformers for Image Recognition at Scale},
	author       = {Dosovitskiy, Alexey and Beyer, Lucas and Kolesnikov, Alexander and Weissenborn, Dirk and Zhai, Xiaohua and Unterthiner, Thomas and Dehghani, Mostafa and Minderer, Matthias and Heigold, Georg and Gelly, Sylvain and Uszkoreit, Jakob and Houlsby, Neil},
	year         = 2021,
	booktitle    = {International Conference on Learning Representations (ICLR)}
}

@inproceedings{touvron2021deit,
	title        = {Training Data-Efficient Image Transformers \& Distillation through Attention ({DeiT})},
	author       = {Touvron, Hugo and Cord, Matthieu and Douze, Matthijs and Massa, Francisco and Sablayrolves, Alexandre and J{\'e}gou, Herv{\'e}},
	year         = 2021,
	booktitle    = {International Conference on Machine Learning (ICML)},
	pages        = {10347--10357}
}

@inproceedings{tan2019efficientnet,
	title        = {{EfficientNet}: Rethinking Model Scaling for Convolutional Neural Networks},
	author       = {Tan, Mingxing and Le, Quoc},
	year         = 2019,
	booktitle    = {International Conference on Machine Learning (ICML)},
	pages        = {6105--6114}
}

@article{landis1977measurement,
	title        = {The Measurement of Observer Agreement for Categorical Data},
	author       = {Landis, J. Richard and Koch, Gary G.},
	year         = 1977,
	journal      = {Biometrics},
	volume       = 33,
	number       = 1,
	pages        = {159--174},
	doi          = {10.2307/2529310}
}

@article{ou2025mmsvpr,
	title        = {{MMS-VPR}: Multimodal Street-Level Visual Place Recognition Dataset and Benchmark},
	author       = {Ou, Yiwei and Ren, Xiaobin and Sun, Ronggui and Gao, Guansong and Zhao, Kaiqi and Manfredini, Manfredo},
	year         = 2025,
	journal      = {arXiv preprint arXiv:2505.12254}
}

@article{bitner1992servicescape,
	title        = {Servicescapes: The Impact of Physical Surroundings on Customers and Employees},
	author       = {Bitner, Mary Jo},
	year         = 1992,
	journal      = {Journal of Marketing},
	volume       = 56,
	number       = 2,
	pages        = {57--71},
	doi          = {10.1177/002224299205600205}
}

@book{goffman1959,
	title        = {The Presentation of Self in Everyday Life},
	author       = {Goffman, Erving},
	year         = 1959,
	publisher    = {Anchor Books},
	address      = {New York}
}

@article{Flickr30k,
	title        = {{Flickr30k entities: Collecting region-to-phrase correspondences for richer image-to-sentence models}},
	author       = {Plummer, Bryan A. and Wang, Liwei and Cervantes, Chris M. and Caicedo, Juan C. and Hockenmaier, Julia and Lazebnik, Svetlana},
	year         = 2015,
	journal      = {Proceedings of the IEEE International Conference on Computer Vision},
	volume       = {2015 International Conference on Computer Vision, ICCV 2015},
	pages        = {2641--2649},
	isbn         = 9781467383912,
	issn         = 15505499
}

@article{ADE20k,
	title        = {{Semantic Understanding of Scenes Through the ADE20K Dataset}},
	author       = {Zhou, Bolei and Zhao, Hang and Puig, Xavier and Xiao, Tete and Fidler, Sanja and Barriuso, Adela and Torralba, Antonio},
	year         = 2019,
	journal      = {International Journal of Computer Vision},
	volume       = 127,
	number       = 3,
	pages        = {302--321},
	issn         = 15731405
}

@article{cityscapes,
	title        = {{Cityscapes}},
	author       = {Alexander, Philip},
	year         = 2022,
	journal      = {Methodist DeBakey cardiovascular journal},
	volume       = 18,
	number       = 2,
	pages        = {114--116},
	issn         = 19476108,
	pmid         = 35414853
}

@article{Lvis,
	title        = {{Lvis: A dataset for large vocabulary instance segmentation}},
	author       = {Gupta, Agrim and Dollar, Piotr and Girshick, Ross},
	year         = 2019,
	journal      = {Proceedings of the IEEE Computer Society Conference on Computer Vision and Pattern Recognition},
	volume       = {2019-June},
	pages        = {5351--5359},
	isbn         = 9781728132938,
	issn         = 10636919
}

@inproceedings{deng2009imagenet,
	title        = {{ImageNet}: A Large-Scale Hierarchical Image Database},
	author       = {Deng, Jia and Dong, Wei and Socher, Richard and Li, Li-Jia and Li, Kai and Fei-Fei, Li},
	year         = 2009,
	booktitle    = {Proceedings of the IEEE Conference on Computer Vision and Pattern Recognition (CVPR)},
	pages        = {248--255},
	doi          = {10.1109/CVPR.2009.5206848}
}

@article{he2025urbanfeel,
	title        = {{UrbanFeel}: A Comprehensive Benchmark for Temporal and Perceptual Understanding of City Scenes through Human Perspective},
	author       = {He, Jun and Lin, Yi and Huang, Zilong and Yin, Jiacong and Ye, Junyan and Zhou, Yuchuan and Li, Weijia and Zhang, Xiang},
	year         = 2025,
	journal      = {arXiv preprint arXiv:2509.22228},
	url          = {https://arxiv.org/abs/2509.22228}
}

@article{kaplan2020scaling,
	title        = {Scaling Laws for Neural Language Models},
	author       = {Kaplan, Jared and McCandlish, Sam and Henighan, Tom and Brown, Tom B. and Chess, Benjamin and Child, Rewon and Gray, Scott and Radford, Alec and Wu, Jeffrey and Amodei, Dario},
	year         = 2020,
	journal      = {arXiv preprint arXiv:2001.08361}
}

@inproceedings{brown2020gpt3,
	title        = {Language Models are Few-Shot Learners},
	author       = {Brown, Tom B. and Mann, Benjamin and Ryder, Nick and Subbiah, Melanie and Kaplan, Jared and Dhariwal, Prafulla and Neelakantan, Arvind and Shyam, Pranav and Sastry, Girish and Askell, Amanda and Agarwal, Sandhini and Herbert-Voss, Ariel and Krueger, Gretchen and Henighan, Tom and Child, Rewon and Ramesh, Aditya and Ziegler, Daniel M. and Wu, Jeffrey and Winter, Clemens and Hesse, Christopher and Chen, Mark and Sigler, Eric and Litwin, Mateusz and Gray, Scott and Chess, Benjamin and Clark, Jack and Berner, Christopher and McCandlish, Sam and Radford, Alec and Sutskever, Ilya and Amodei, Dario},
	year         = 2020,
	booktitle    = {Advances in Neural Information Processing Systems (NeurIPS)},
	volume       = 33,
	pages        = {1877--1901}
}

@article{anguelov2010streetview,
	title={Google street view: Capturing the world at street level},
	author={Anguelov, Dragomir and Dulong, Carole and Filip, Daniel and Frueh, Christian and Lafon, St{\'e}phane and Lyon, Richard and Ogale, Abhijit and Vincent, Luc and Weaver, Josh},
	journal={Computer},
	volume={43},
	number={6},
	pages={32--38},
	year={2010},
	publisher={IEEE}
}

@book{newman1972,
	title        = {Defensible Space: Crime Prevention through Urban Design},
	author       = {Newman, Oscar},
	year         = 1972,
	publisher    = {Macmillan},
	address      = {New York}
}

@article{peredy2024chinese,
	title={Chinese city tier ranking scheme as special spatial factor of innovations diffusion},
	author={Peredy, Zolt{\'a}n and Li, Sijia and V{\'\i}gh, L{\'a}szl{\'o}},
	journal={International Review},
	number={1-2},
	pages={88--99},
	year={2024}
}

\clearpage
\appendix


\section{Full Experimental Results}
\label{app:full_results}

\subsection{Task~1: Per-Class Diagnostics}
\label{app:t1_full}

\begin{figure}[h]
  \centering
  \includegraphics[width=\linewidth]{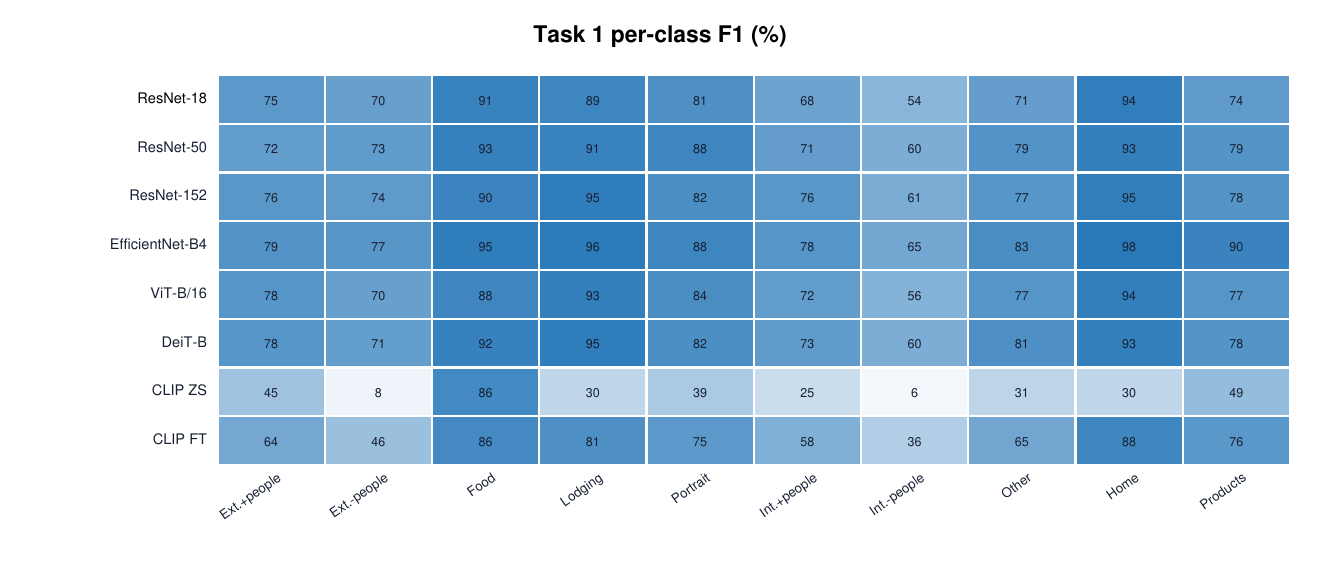}
  \caption{Task~1 per-class F1 for all baselines.
  Classes~3, 4, 5 are consistently the most challenging;
  the activated/non-activated boundary pairs (0/1, 2/3) show systematic
  confusion across all models.}
  \label{fig:t1_per_class}
\end{figure}

\begin{figure}[h]
  \centering
  \includegraphics[width=0.78\linewidth]{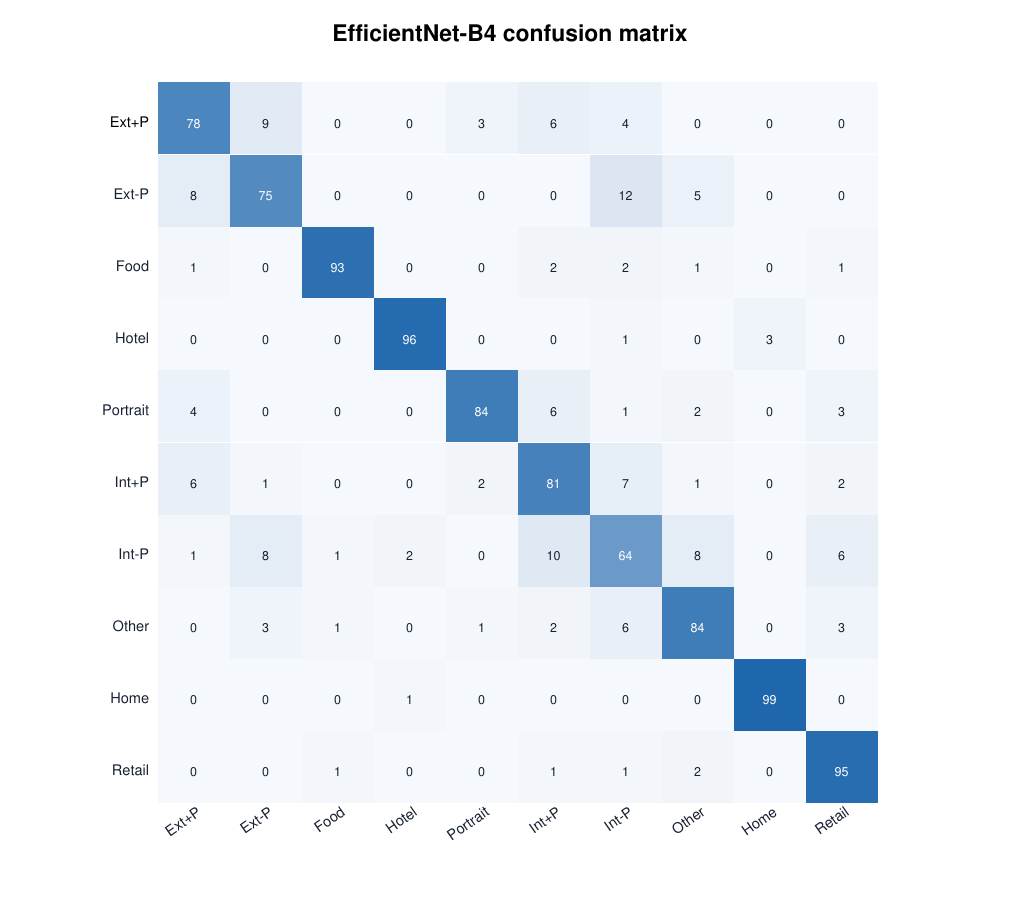}
  \caption{Confusion matrix for EfficientNet-B4.
  Off-diagonal mass concentrates on the activated/non-activated boundary
  pairs and the commercial lodging vs.\ private residential distinction.}
  \label{fig:t1_confusion}
\end{figure}

Figure~\ref{fig:t1_per_class} shows per-class F1 scores for all T1 baselines
on the 100K tier (10K held-out test split).
The most challenging categories are \textit{Interior without People}
(Class~3) and the activated/non-activated boundary pairs (Classes~0/1 and
2/3), where the sole distinguishing cue is the presence or absence of people.
Accommodation classes (4 and 5) are similarly difficult due to visual overlap
between hotel and residential interiors at lower training scales.
Figure~\ref{fig:t1_confusion} confirms that EfficientNet-B4's off-diagonal
confusion concentrates precisely on these boundary pairs.

\subsection{Task~2: Direction-Level Retrieval Results}
\label{app:t2_full}

Table~\ref{tab:t2_direction} reports Task~2 results broken down by retrieval
direction (T2I: text$\to$image; I2T: image$\to$text) for all model and
setting combinations on the 10K test split, including BLIP-2 results
discussed in Section~\ref{sec:discussion}.
Category-label queries are symmetric by construction (10 fixed text prompts
each matched against all images of that class), so T2I and I2T values are
identical in that setting.
Post-text retrieval shows a mild asymmetry: I2T slightly outperforms T2I
because the image-to-text candidate pool (943 unique post texts) is marginally
smaller than the image gallery (1,000 images), reducing the retrieval burden.

\begin{table*}[h]
\centering
\caption{Task~2 retrieval results by direction.
  T2I\,=\,text$\to$image; I2T\,=\,image$\to$text.
  ZS\,=\,zero-shot; FT\,=\,fine-tuned. MedR: lower is better.
  mAP values reported as percentages (0--100).
  Category-label queries are symmetric; T2I and I2T values are identical.}
\label{tab:t2_direction}
\setlength{\tabcolsep}{3.5pt}
\renewcommand{\arraystretch}{1.10}
\small
\begin{tabular}{llllccccc}
\toprule
\textbf{Setting} & \textbf{Model} & \textbf{Mode} & \textbf{Dir.} &
\textbf{R@1} & \textbf{R@5} & \textbf{R@10} & \textbf{mAP} & \textbf{MedR} \\
\midrule
\multirow{12}{*}{Category label}
  & \multirow{4}{*}{CLIP~\citep{radford2021clip}}
      & \multirow{2}{*}{ZS} & T2I & 54.2 & 96.5 & 100.0 & 53.3 & 1.5 \\
  &   &                     & I2T & 54.2 & 96.5 & 100.0 & 53.3 & 1.5 \\
  &   & \multirow{2}{*}{FT} & T2I & 92.7 & 99.8 & 100.0 & 90.7 & 1.0 \\
  &   &                     & I2T & 92.7 & 99.8 & 100.0 & 90.7 & 1.0 \\
\cmidrule(l){2-9}
  & \multirow{4}{*}{BLIP~\citep{li2022blip}}
      & \multirow{2}{*}{ZS} & T2I & 14.9 & 43.6 &  80.0 & 19.8 & 6.2 \\
  &   &                     & I2T & 14.9 & 43.6 &  80.0 & 19.8 & 6.2 \\
  &   & \multirow{2}{*}{FT} & T2I & \textbf{94.2} & \textbf{99.8} & \textbf{100.0} & \textbf{93.3} & \textbf{1.0} \\
  &   &                     & I2T & \textbf{94.2} & \textbf{99.8} & \textbf{100.0} & \textbf{93.3} & \textbf{1.0} \\
\cmidrule(l){2-9}
  & \multirow{4}{*}{BLIP-2~\citep{li2023blip2}}
      & \multirow{2}{*}{ZS} & T2I & 34.6 & 70.0 &  90.0 & 36.6 & 2.5 \\
  &   &                     & I2T & 34.6 & 70.0 &  90.0 & 36.6 & 2.5 \\
  &   & \multirow{2}{*}{FT} & T2I & 93.4 & 99.8 & 100.0 & 92.0 & 1.0 \\
  &   &                     & I2T & 93.4 & 99.8 & 100.0 & 92.0 & 1.0 \\
\midrule
\multirow{12}{*}{Post text}
  & \multirow{4}{*}{CLIP~\citep{radford2021clip}}
      & \multirow{2}{*}{ZS} & T2I &  1.5 &  3.4 &  5.0 &  2.9 & 364 \\
  &   &                     & I2T &  3.7 &  7.4 &  9.1 &  6.1 & 292 \\
  &   & \multirow{2}{*}{FT} & T2I &  \textbf{7.9} & \textbf{16.3} & \textbf{23.3} & \textbf{12.7} & \textbf{63} \\
  &   &                     & I2T &  \textbf{8.2} & \textbf{17.4} & \textbf{23.6} & \textbf{13.7} & \textbf{64} \\
\cmidrule(l){2-9}
  & \multirow{4}{*}{BLIP~\citep{li2022blip}}
      & \multirow{2}{*}{ZS} & T2I &  0.2 &  0.5 &  1.1 &  0.8 & 490 \\
  &   &                     & I2T &  0.0 &  0.2 &  1.4 &  0.7 & 464 \\
  &   & \multirow{2}{*}{FT} & T2I &  1.4 &  6.5 & 10.4 &  4.8 &  95 \\
  &   &                     & I2T &  2.5 &  7.2 & 12.8 &  6.3 &  89 \\
\cmidrule(l){2-9}
  & \multirow{4}{*}{BLIP-2~\citep{li2023blip2}}
      & \multirow{2}{*}{ZS} & T2I &  0.8 &  2.0 &  3.0 &  1.9 & 427 \\
  &   &                     & I2T &  1.9 &  3.8 &  5.3 &  3.4 & 378 \\
  &   & \multirow{2}{*}{FT} & T2I &  4.7 & 11.4 & 16.9 &  8.8 &  79 \\
  &   &                     & I2T &  5.4 & 12.3 & 18.2 & 10.0 &  77 \\
\bottomrule
\end{tabular}
\end{table*}

\subsection{Task~2 Metadata Schema}
\label{app:t2schema}

Table~\ref{tab:t2schema} lists all 15 metadata columns accompanying each
image record in the \texttt{train/val/test.xlsx} files.

\begin{table}[h]
	\centering
	\caption{T2 metadata schema (\texttt{train/val/test.xlsx}).
		\texttt{Image~Filename} is the primary key linking rows to image files.
		All text fields retain original Chinese; column names are English.}
	\label{tab:t2schema}
	\setlength{\tabcolsep}{4pt}
	\renewcommand{\arraystretch}{1.05}
	\small
	\begin{tabular}{l l p{7.5cm}}
		\toprule
		\textbf{Column} & \textbf{Type} & \textbf{Description} \\
		\midrule
		\texttt{Image Label}         & Categorical  &
		\husic{} class name; T1 ground truth; T2-1 query text \\
		\texttt{Image Filename}      & String       &
		\texttt{UserID\_PostTime\_Index}; primary join key to image files \\
		\texttt{Post ID}             & Integer      &
		Encrypted numeric post identifier \\
		\texttt{User ID}             & Integer      &
		Cryptographically encrypted user code \\
		\texttt{Post Time}           & DateTime     &
		Publication timestamp \\
		\texttt{Post Text}           & String (ZH)  &
		Full original post text; T2-2 query/target \\
		\texttt{City}                & Categorical  &
		City of venue \\
		\texttt{Place Tag}           & String       &
		Venue hashtag \\
		\texttt{Posting Tool}        & String       &
		Client device / platform \\
		\texttt{Mentioned Users}     & List[Int]    &
		Encrypted mentioned user IDs \\
		\texttt{Extracted Topics}    & List[String] &
		Hashtag topics \\
		\texttt{Extracted Locations} & List[String] &
		Additional geo-tags \\
		\texttt{Like Count}          & Integer      &
		Total likes \\
		\texttt{Repost Count}        & Integer      &
		Total reposts \\
		\texttt{Comment Count}       & Integer      &
		Total comments \\
		\bottomrule
	\end{tabular}
\end{table}

\textbf{Join protocol.}
Images are matched to metadata rows via \texttt{Image~Filename}.
Multiple rows sharing the same \texttt{Post~ID} share one
\texttt{Post~Text}; grouping by \texttt{Post~ID} recovers all images
belonging to a single post.

\textbf{T2-1 (category-label) query.}
Use \texttt{Image~Label} as the text query, formatted as
``\texttt{This is a photo of \{class\_name\}}'';
evaluate against all images sharing the same label
(\emph{many-relevant} retrieval, 10 fixed queries).

\textbf{T2-2 (post-text) query.}
Use \texttt{Post~Text} as the text query; the positive image set is all
rows sharing the same \texttt{Post~ID}.
Because one post may include up to nine images, this defines a
\emph{multi-positive} retrieval problem
(see Appendix~\ref{app:t2code} for the evaluation protocol).

\subsection{Task~3: Additional Qualitative Segmentation Results}
\label{app:t3_qualitative}
\begin{figure*}[h]
  \centering
  \begin{subfigure}[b]{0.48\linewidth}
    \includegraphics[width=\linewidth]{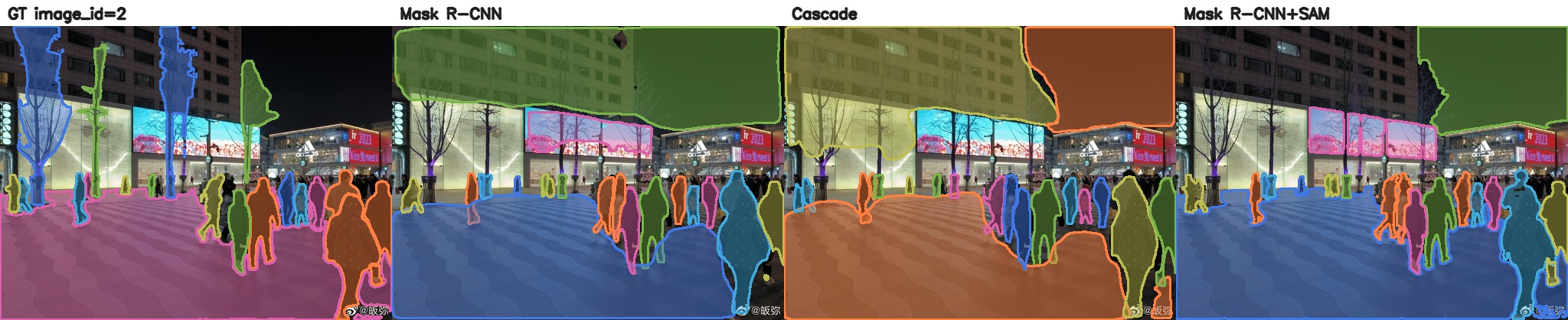}
    \caption{Sample 1}
  \end{subfigure}\hfill
  \begin{subfigure}[b]{0.48\linewidth}
    \includegraphics[width=\linewidth]{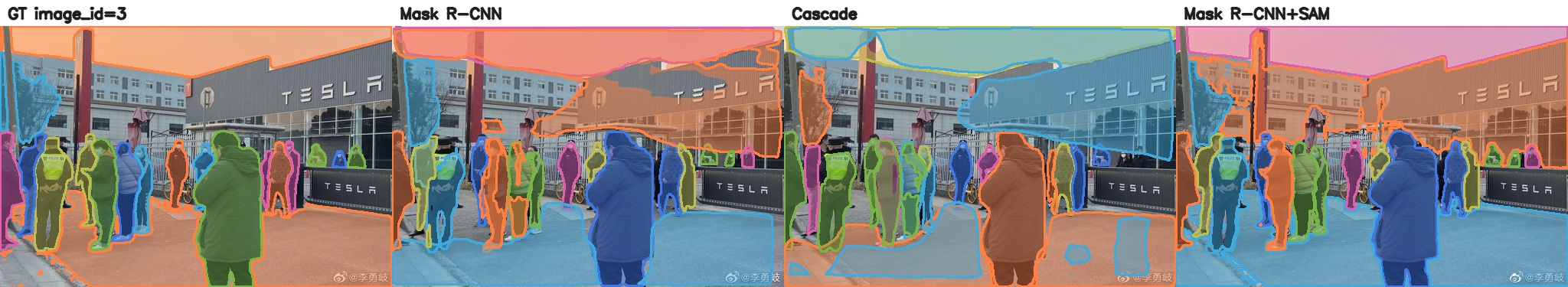}
    \caption{Sample 2}
  \end{subfigure}
  \vspace{4pt}
  \begin{subfigure}[b]{0.48\linewidth}
    \includegraphics[width=\linewidth]{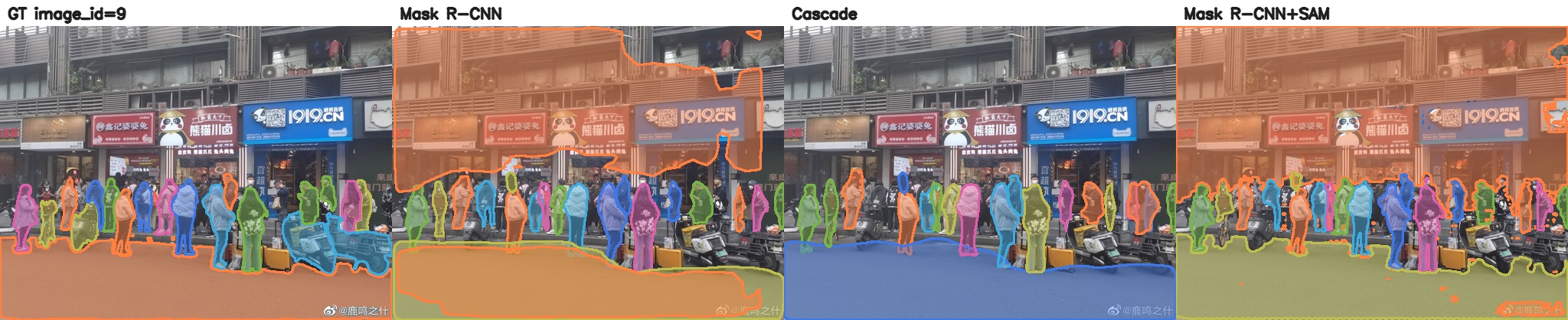}
    \caption{Sample 3}
  \end{subfigure}\hfill
  \begin{subfigure}[b]{0.48\linewidth}
    \includegraphics[width=\linewidth]{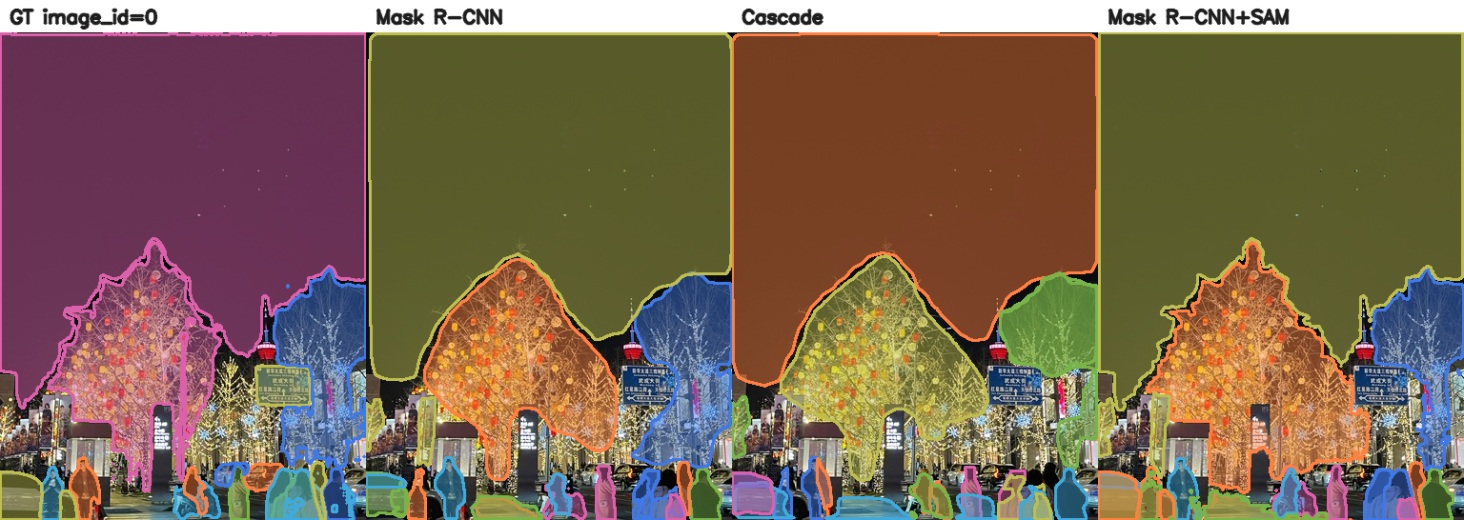}
    \caption{Sample 4}
  \end{subfigure}
  \vspace{4pt}
  \begin{subfigure}[b]{0.48\linewidth}
    \includegraphics[width=\linewidth]{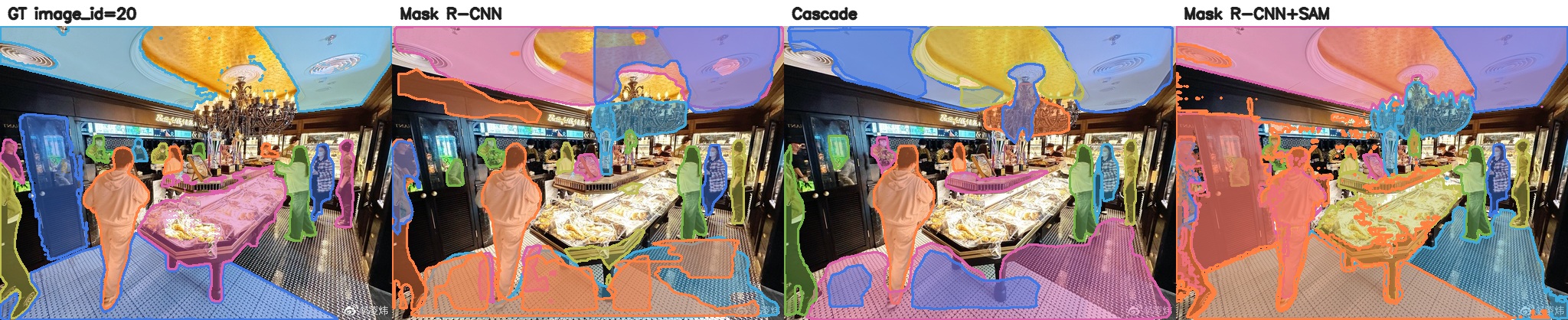}
    \caption{Sample 5}
  \end{subfigure}\hfill
  \begin{subfigure}[b]{0.48\linewidth}
    \includegraphics[width=\linewidth]{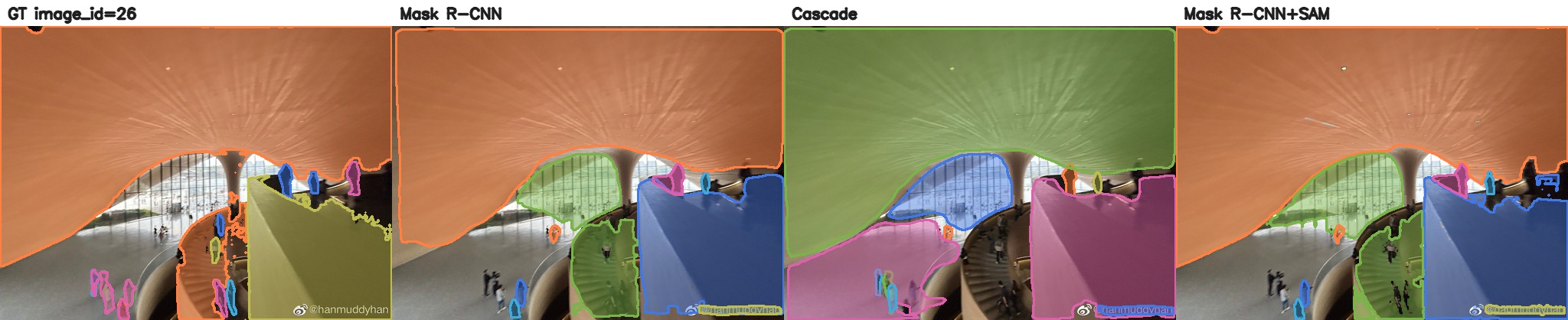}
    \caption{Sample 6}
  \end{subfigure}
  \caption{Task~3 qualitative results (Samples 1--6).
  Left: ground-truth pseudo-label overlay;
  right: Cascade Mask~R-CNN\,+\,SAM predictions.}
  \label{fig:qual_all}
\end{figure*}

Figure~\ref{fig:qual_all} presents additional qualitative
instance segmentation results, complementing the main-body panel
(Figure~\ref{fig:seg_examples}).
Each panel shows the ground-truth pseudo-label overlay alongside
Cascade Mask~R-CNN\,+\,SAM predictions.
The per-class prompt vocabulary (Appendix~\ref{app:segvocab}) enables
detection of domain-specific objects---escalators, retail shelves, hotel
furnishings, outdoor architectural elements---absent from general-purpose
segmentation vocabularies.
SAM box-refinement consistently produces tighter, more complete masks than
the standalone detector, particularly for glass surfaces and architectural
structures with diffuse boundaries.

\section{Scaling Behaviour: Full Results}
\label{app:scaling}

To study domain-specific scaling, we trained all models separately on
balanced 1K, 10K, and 100K tiers and evaluated each on a shared held-out
10K test set.
Tables~\ref{tab:scaling_t1} and~\ref{tab:scaling_hier} report T1 accuracy
and macro-F1; Table~\ref{tab:t2_scaling} reports T2-Post retrieval scaling.

\textbf{T1 scaling.}
All models improve monotonically with scale:
ResNet-50 progresses 66.5\% $\to$ 78.1\% $\to$ 83.5\%;
ResNet-152 from 67.3\% $\to$ 79.0\% $\to$ 83.5\%;
CLIP fine-tuned from 70.8\% $\to$ 78.0\% $\to$ 82.3\%.
LLaVA-1.5 fine-tuned achieves the highest accuracy at small scales
(76.8\% at 1K, 81.2\% at 10K), confirming its strong language-grounded
priors, but is computationally prohibitive at 100K scale
($\approx$3,200$\times$ slower per sample than ResNet-50; see
Appendix~\ref{app:cost}).
The 1K$\to$10K improvement (10--12\,pp) consistently exceeds the
10K$\to$100K gain ($\approx$5\,pp), consistent with standard scaling
laws~\citep{kaplan2020scaling}.

\textbf{Hierarchical granularity.}
At the 100K tier, models achieve significantly higher accuracy on coarser
\husic{} distinctions (spatial vs.\ non-spatial: $\approx$94\%;
exterior vs.\ interior: $\approx$95\%) than on the fine-grained 10-class
task ($\approx$83--85\%), confirming that \husic{} captures semantically
meaningful hierarchical structure.

\textbf{T2-Post retrieval scaling.}
CLIP fine-tuned average R@1 drops from 39.5\% on the 1K split
(100-image retrieval pool) to 8.1\% on the 10K split (1,000-image pool),
confirming that post-level retrieval is a scalably challenging benchmark
as the candidate gallery grows.

\begin{table*}[h]
\centering
\caption{T1 scaling: fine-tuned top-1 accuracy and macro-F1 on the shared
  10K held-out test set across three training tiers.
  LLaVA 100K fine-tuning not completed due to computational constraints.}
\label{tab:scaling_t1}
\setlength{\tabcolsep}{5pt}
\renewcommand{\arraystretch}{1.08}
\small
\begin{tabular}{l cc cc cc}
\toprule
& \multicolumn{2}{c}{\textbf{1K tier}} &
  \multicolumn{2}{c}{\textbf{10K tier}} &
  \multicolumn{2}{c}{\textbf{100K tier}} \\
\cmidrule(lr){2-3}\cmidrule(lr){4-5}\cmidrule(lr){6-7}
\textbf{Model} & Acc.\ (\%) & F1 & Acc.\ (\%) & F1 & Acc.\ (\%) & F1 \\
\midrule
ResNet-50~\citep{he2016resnet}  & 66.5 & 0.661 & 78.1 & 0.781 & 83.5 & 0.835 \\
ResNet-152~\citep{he2016resnet} & 67.3 & 0.670 & 79.0 & 0.787 & 83.5 & 0.834 \\
CLIP (FT)~\citep{radford2021clip} & 70.8 & 0.708 & 78.0 & 0.780 & 82.3 & 0.822 \\
LLaVA-1.5 (FT)~\citep{liu2024llava} & 76.8 & 0.767 & 81.2 & 0.812 & --- & --- \\
\bottomrule
\end{tabular}
\end{table*}

\begin{table*}[h]
\centering
\caption{Hierarchical T1 scaling: accuracy at three \husic{} granularity
  levels on the shared 10K held-out test set (fine-tuned models).}
\label{tab:scaling_hier}
\setlength{\tabcolsep}{4pt}
\renewcommand{\arraystretch}{1.08}
\small
\begin{tabular}{l c cc cc cc}
\toprule
& & \multicolumn{2}{c}{\textbf{Spatial / Non-spatial}} &
  \multicolumn{2}{c}{\textbf{Exterior / Interior}} &
  \multicolumn{2}{c}{\textbf{10-class}} \\
\cmidrule(lr){3-4}\cmidrule(lr){5-6}\cmidrule(lr){7-8}
\textbf{Model} & \textbf{Tier} & Acc.\ (\%) & F1 & Acc.\ (\%) & F1 & Acc.\ (\%) & F1 \\
\midrule
\multirow{3}{*}{ResNet-50}
 & 1K   & 88.7 & 0.884 & 86.7 & 0.867 & 66.5 & 0.661 \\
 & 10K  & 92.5 & 0.922 & 92.3 & 0.923 & 78.1 & 0.781 \\
 & 100K & 93.9 & 0.937 & 95.0 & 0.950 & 83.5 & 0.835 \\
\midrule
\multirow{3}{*}{ResNet-152}
 & 1K   & 90.0 & 0.896 & 85.8 & 0.858 & 67.3 & 0.670 \\
 & 10K  & 92.9 & 0.926 & 92.1 & 0.920 & 79.0 & 0.787 \\
 & 100K & 94.2 & 0.939 & 94.7 & 0.947 & 83.5 & 0.834 \\
\midrule
\multirow{3}{*}{CLIP (FT)}
 & 1K   & 90.1 & 0.898 & 76.8 & 0.765 & 70.8 & 0.708 \\
 & 10K  & 92.8 & 0.925 & 84.9 & 0.849 & 78.0 & 0.780 \\
 & 100K & 94.0 & 0.937 & 87.5 & 0.875 & 82.3 & 0.822 \\
\midrule
\multirow{2}{*}{LLaVA-1.5 (FT)}
 & 1K   & 90.6 & 0.905 & 80.5 & 0.807 & 76.8 & 0.767 \\
 & 10K  & 91.9 & 0.930 & 85.4 & 0.865 & 81.2 & 0.812 \\
\bottomrule
\end{tabular}
\end{table*}

\begin{table*}[h]
\centering
\caption{T2-Post retrieval scaling: average R@1 and mAP for fine-tuned
  models across 1K and 10K evaluation splits.
  The retrieval pool grows from 100 images (1K) to 1,000 images (10K),
  naturally reducing R@$K$ scores.
  Average of T2I and I2T directions; mAP as a fraction (0--1).}
\label{tab:t2_scaling}
\setlength{\tabcolsep}{5pt}
\renewcommand{\arraystretch}{1.08}
\small
\begin{tabular}{l cc cc}
\toprule
& \multicolumn{2}{c}{\textbf{1K split (100-img pool)}} &
  \multicolumn{2}{c}{\textbf{10K split (1,000-img pool)}} \\
\cmidrule(lr){2-3}\cmidrule(lr){4-5}
\textbf{Model} & Avg.\ R@1 (\%) & Avg.\ mAP & Avg.\ R@1 (\%) & Avg.\ mAP \\
\midrule
CLIP (FT)~\citep{radford2021clip}     & 39.5 & 0.501 & 8.1 & 0.132 \\
BLIP-2 (FT)~\citep{li2023blip2}       & 28.1 & 0.392 & 5.0 & 0.094 \\
BLIP (FT)~\citep{li2022blip}          & 16.6 & 0.283 & 1.9 & 0.055 \\
\bottomrule
\end{tabular}
\end{table*}

\section{Computational Cost}
\label{app:cost}

Table~\ref{tab:cost} summarises hardware, training time, inference speed,
and accuracy for representative model--tier combinations.
LLaVA-1.5 100K fine-tuning was not completed due to its estimated training
time exceeding $\approx$150 GPU-hours on H100 (per-sample cost
$\approx$3,200$\times$ that of ResNet-50).

\begin{table*}[h]
	\centering
	\caption{Training and inference costs across dataset scales for T1 models.
		Accuracy evaluated on the shared held-out 10K test set.
		Inference time measured per image on the respective device.}
	\label{tab:cost}
	\setlength{\tabcolsep}{4pt}
	\renewcommand{\arraystretch}{1.10}
	\small
	\begin{tabular}{llllrrc}
		\toprule
		\textbf{Scale} & \textbf{Model} & \textbf{Hardware} &
		\textbf{Train (s)} & \textbf{Inf.\ (s/img)} & \textbf{Top-1 Acc.\ (\%)} \\
		\midrule
		\multirow{4}{*}{1K}
		& ResNet-50~\citep{he2016resnet}   & CPU (M2 Pro)      &    291 & 0.013 & 66.5 \\
		& ResNet-152~\citep{he2016resnet}  & CPU (M2 Pro)      &    568 & 0.011 & 67.3 \\
		& CLIP~\citep{radford2021clip}     & A100-80G          &    210 & 0.005 & 70.8 \\
		& LLaVA-1.5~\citep{liu2024llava}  & A100-80G          &  2{,}121 & 0.498 & 76.8 \\
		\midrule
		\multirow{4}{*}{10K}
		& ResNet-50                        & M2 Pro            &  2{,}709 & 0.005 & 78.1 \\
		& ResNet-152                       & M2 Pro            &  5{,}751 & 0.008 & 79.0 \\
		& CLIP                             & H100-80G          &  1{,}460 & 0.003 & 78.0 \\
		& LLaVA-1.5                        & H100-80G          & 11{,}565 & 1.256 & 81.2 \\
		\midrule
		\multirow{4}{*}{100K}
		& ResNet-50                        & A100-40G          &  2{,}435 & 0.001 & 83.5 \\
		& ResNet-152                       & A100-40G          &  4{,}609 & 0.001 & 83.5 \\
		& CLIP                             & H100-80G          & 11{,}606 & 0.003 & 82.3 \\
		& LLaVA-1.5                        & H100-80G          & \multicolumn{3}{l}{Not available (resource constraints)} \\
		\bottomrule
	\end{tabular}
\end{table*}

\section{Task~2 Retrieval Evaluation Protocol}
\label{app:t2code}

\textbf{Multi-positive matching protocol.}
The T2-Post evaluation uses many-to-many positive matching.
For T2I (text$\to$image): each unique post text is a query; all images from
that post constitute the positive set.
For I2T (image$\to$text): each image is a query; its corresponding post text
is the single positive.
A query is a hit at rank $K$ if \emph{any one} of its positive targets
appears within the top-$K$ retrieved results.

\textbf{Random-chance baselines (10K split).}
T2I: $\approx$0.106\% R@1 (1.06 positives per query / 1,000-image gallery);
I2T: $\approx$0.106\% R@1 (1 positive per image / 943 unique post texts).
Fine-tuned CLIP achieves average R@1\,=\,8.1\%, approximately $76\times$
random chance.

\textbf{Quick-start.}
The full evaluation script (\texttt{run\_task2\_multipositive.py}) is
released in \ourname{}-lib.
Key flags: \texttt{--text-source~label} (T2-1, category prompts) or
\texttt{--text-source~post} (T2-2, original post text);
\texttt{--do-finetune} enables fine-tuning before evaluation.
Minimum RAM: 8\,GB (1K tier), 32\,GB (10K tier).

\section{Hyperparameter Details}
\label{app:hyperparams}

\textbf{Task~1 (Scene Classification).}
All supervised models are initialised from ImageNet-pretrained weights.
CNN-based models (ResNet-\{18/50/152\}, EfficientNet-B4): AdamW,
LR\,=\,$10^{-4}$, weight decay\,=\,$10^{-4}$, batch size 64, cosine
annealing, 50 epochs with early stopping (patience\,=\,5).
Transformer-based models (ViT-B/16, DeiT-B): AdamW, LR\,=\,$10^{-5}$,
batch size 32, cosine annealing, 50 epochs.
CLIP ViT-L/14 fine-tuning: AdamW, LR\,=\,$5\times10^{-5}$, batch size 32,
50 epochs, text encoder frozen.
LLaVA-1.5 fine-tuning: LoRA ($r\,=\,16$), AdamW,
LR\,=\,$2\times10^{-4}$, 3 epochs.
Data augmentation: random horizontal flip, colour jitter, random crop;
input resolution $224\times224$, ImageNet normalisation.

\textbf{Task~2 (Cross-Modal Retrieval).}
CLIP fine-tuning: InfoNCE contrastive loss, temperature $\tau\,=\,0.07$,
LR\,=\,$10^{-6}$, batch size 32, AdamW, 3 epochs; both T2I and I2T
directions trained jointly.
BLIP fine-tuning: ITM\,+\,ITC joint loss, LR\,=\,$10^{-5}$, batch size 16,
3 epochs.
BLIP-2 fine-tuning: LoRA adaptation ($r\,=\,16$), LR\,=\,$2\times10^{-5}$,
batch size 16, 3 epochs.
All retrieval models are evaluated under the multi-positive protocol
described in Appendix~\ref{app:t2code}.
Post-group membership is determined from the \texttt{Post~ID} field in the
metadata spreadsheet (\texttt{train/val/test.xlsx}).

\textbf{Task~3 (Instance Segmentation).}
Mask~R-CNN and Cascade Mask~R-CNN: ResNet-50 backbone with FPN,
initialised from COCO-pretrained weights via MMDetection.
SGD, LR\,=\,0.02, 8 epochs, batch size 2, multi-scale training
(640--800 short side), class-agnostic head (single foreground category).
SAM box-refinement: bounding-box proposals from the paired detector
(score threshold 0.001) are passed to SAM ViT-H as box prompts;
no SAM fine-tuning is performed.
GT-box SAM oracle: SAM ViT-H with ground-truth bounding boxes; no training.
All T3 models trained on $\approx$45K pseudo-label annotations
(Grounding~DINO confidence $\ge\,0.35$, SAM~2 predicted IoU $\ge\,0.80$).

\section{Extended Discussion and Limitations}
\label{app:discussion}
\begin{enumerate}[leftmargin=1.8em,itemsep=2pt,topsep=4pt]
  \item \textbf{Class-agnostic T3 evaluation.}
        Instance segmentation is benchmarked under a single
        \texttt{object} category; per-class breakdown would require
        higher-quality human-annotated ground truth than current
        pseudo-labels support.

  \item \textbf{Geographic restriction to Chinese cities.}
        All 61 venues are located across 24 Chinese cities; whether the
        \husic{} taxonomy and learned representations generalise to other
        cultural or urban contexts requires future geographic expansion.

  \item \textbf{Class imbalance in the 2M corpus.}
        The full 2M corpus is class-imbalanced by construction, reflecting
        real-world social media frequency distributions (non-spatial classes
        each comprising $\approx$15--25\% of posts; all spatially relevant
        classes collectively $\approx$40\%).
        Researchers requiring balanced training at scale should use the
        100K tier.

  \item \textbf{Incomplete LLaVA-1.5 100K training.}
        100K fine-tuning of LLaVA-1.5 was not completed due to
        computational constraints; 1K and 10K results are reported but
        100K results are unavailable.

  \item \textbf{T3 SAM oracle circularity.}
        The GT-box SAM oracle (AP\,=\,0.749) partially reflects
        circularity, as evaluation pseudo-labels were generated by SAM
        and the oracle uses SAM with perfect box prompts.
        The Cascade Mask~R-CNN and SAM box-refinement results---trained
        on noisy pseudo-labels and evaluated against stricter-threshold
        human-audited annotations---provide the more informative measure
        of generalisable performance.

  \item \textbf{Chinese-language social media text.}
        Post-text retrieval operates on original Chinese Weibo posts.
        Current baselines (CLIP, BLIP, BLIP-2) were pre-trained
        predominantly on English data, which partly explains the low
        absolute post-level retrieval scores and motivates future
        bilingual or multilingual urban-domain pre-training.
\end{enumerate}


\section{The HUSIC 10-Class Framework}
\label{app:husic}

Table~\ref{tab:husic} presents the full \husic{} taxonomy, including
formal definitions, theoretical grounding, and downstream research value
for each class.
Classes~0--5 (Spatially Relevant) receive T3 instance segmentation
pseudo-labels; Classes~6--9 (Non-Spatially Relevant) support T1
classification and serve as the UGC filtering baseline.

\begin{table*}[h]
\centering
\caption{Full \husic{} framework with definitions, theoretical grounding,
  and research significance.}
\label{tab:husic}
\setlength{\tabcolsep}{4pt}
\renewcommand{\arraystretch}{1.10}
\small
\begin{tabular}{c >{\raggedright\arraybackslash}p{2.2cm}
  >{\raggedright\arraybackslash}p{3.0cm}
  >{\raggedright\arraybackslash}p{3.3cm}
  >{\raggedright\arraybackslash}p{3.8cm}}
\toprule
\textbf{ID} & \textbf{Label} & \textbf{Definition} &
\textbf{Theoretical grounding} & \textbf{Research value} \\
\midrule
\multicolumn{5}{l}{\textit{Spatially Relevant --- Urban Exterior}} \\
0 & Exterior urban spaces with people
  & Outdoor urban spaces with human presence and activity
  & Gehl: social/optional activity~\citep{gehl2011};
    Space Syntax~\citep{hillier1984}
  & Spatial vitality; pedestrian behaviour; crowding analysis \\[2pt]
1 & Exterior urban spaces without people
  & Outdoor architectural elements without human figures
  & Lefebvre: conceived space~\citep{lefebvre1991};
    Lynch: imageability~\citep{lynch1960}
  & Architectural attractiveness; design quality \\[2pt]
\midrule
\multicolumn{5}{l}{\textit{Spatially Relevant --- Urban Public Interior}} \\
2 & Interior urban spaces with people
  & Indoor commercial spaces with visible occupants
  & Newman: spatial hierarchy and publicity gradient~\citep{newman1972}
  & Indoor retail vitality; commercial activation \\[2pt]
3 & Interior urban spaces without people
  & Indoor spaces showing design elements only
  & Bitner: servicescape~\citep{bitner1992servicescape}
  & Interior design quality; aesthetic perception \\[2pt]
\midrule
\multicolumn{5}{l}{\textit{Spatially Relevant --- Accommodation}} \\
4 & Hotel or commercial lodging spaces
  & Hotel and serviced apartment interiors
  & Privacy regulation theory; hospitality servicescape
  & Tourism infrastructure; commercial--residential interaction \\[2pt]
5 & Private home interiors
  & Private home interiors near commercial hubs
  & Urban proximity effects; housing market studies
  & Commercial influence on local rental market \\[2pt]
\midrule
\multicolumn{5}{l}{\textit{Non-Spatially Relevant}} \\
6 & Food or drink items
  & Food, beverage, and dining photography
  & Servicescape~\citep{bitner1992servicescape};
    social eating sociology
  & F\&B consumption behaviour; dining space programming \\[2pt]
7 & Retail products and merchandise
  & Merchandise and product display photography
  & Visual merchandising; retail consumer behaviour
  & Consumer preference; product display strategy \\[2pt]
8 & Human-centered portrait
  & Selfies, group photos, and portraits
  & Goffman: performance theory~\citep{goffman1959}
  & Social behaviour documentation; action recognition foundation \\[2pt]
9 & Other non-spatial content
  & Advertisements, screenshots, memes, QR codes
  & Signal-to-noise theory; data quality frameworks
  & UGC filtering baseline; dataset quality assessment \\[2pt]
\bottomrule
\end{tabular}
\end{table*}

\section{Geographic Scope and Site Details}
\label{app:geo}

\begin{figure}[h]
    \centering
    \includegraphics[width=\textwidth]{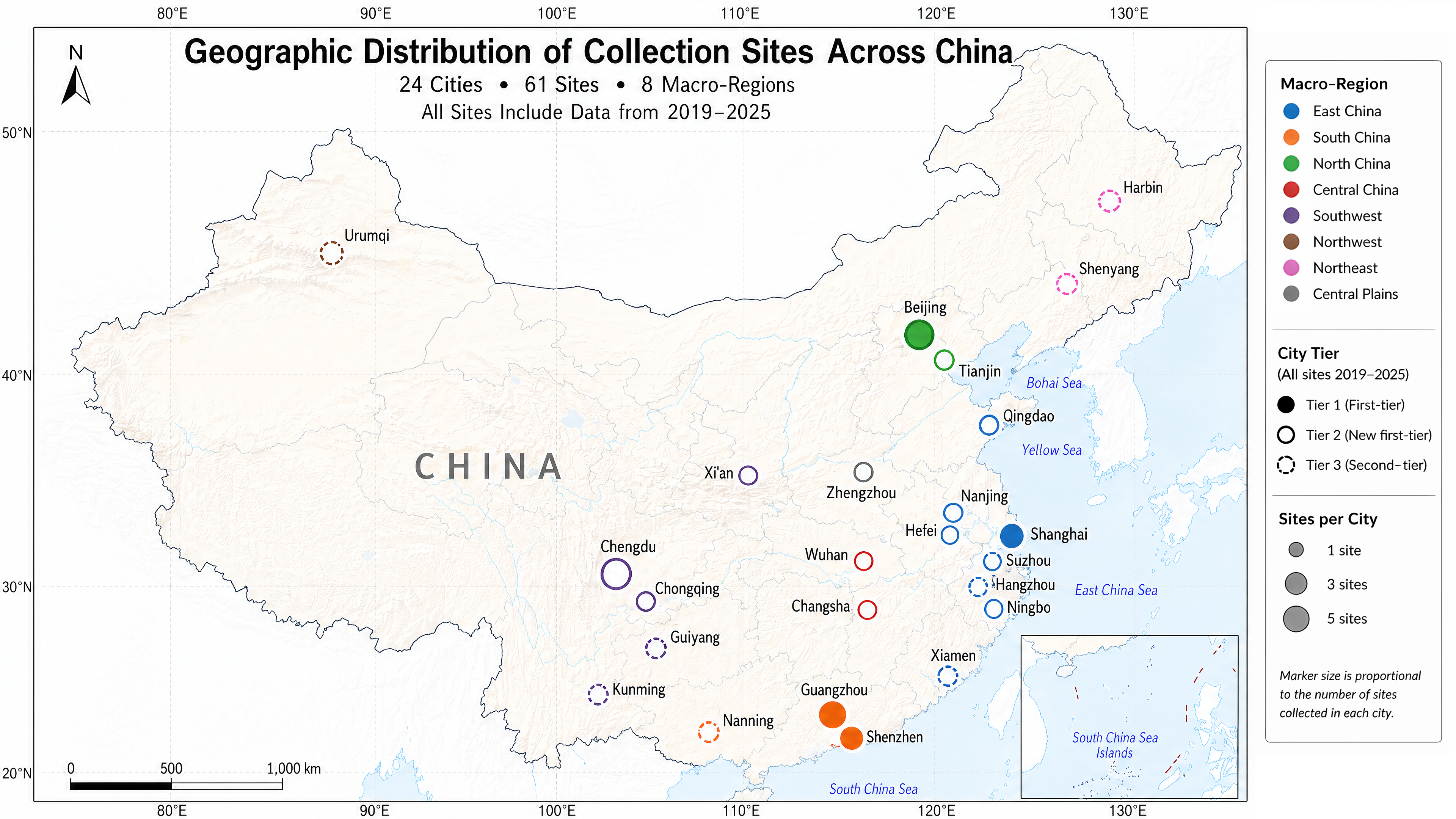}
    \caption{Geographic distribution of \ourname{}'s 24 collection cities.
    Marker size is proportional to the number of collected image--text pairs
    per city; colour encodes macro-region.}
    \label{fig:geo}
\end{figure}

Table~\ref{tab:sites} lists all 61 collection sites across the 24 cities,
organised by macro-region, together with their spatial typology and city-tier
classification.
Figure~\ref{fig:geo} shows the geographic distribution of collection cities.
Sites sharing a brand identity across cities (marked~$\dagger$) enable
controlled same-brand, different-city comparisons unique to \ourname{}.

\begin{table*}[t]
    \centering
    \caption{%
        Geographic scope of \ourname{}: 24 cities and 61 commercial sites,
        organised by macro-region.
        \textbf{Type}: \textbf{E}\,=\,enclosed mall;
        \textbf{O}\,=\,open-air pedestrian precinct;
        \textbf{M}\,=\,mixed-typology.
        \textbf{Tier}: 1\,=\,first-tier (Beijing, Shanghai, Guangzhou,
        Shenzhen); 2\,=\,new first-tier; 3\,=\,second-tier.
        $\dagger$\,=\,brand identity shared across cities.
    }
    \label{tab:sites}
    \setlength{\tabcolsep}{4pt}
    \renewcommand{\arraystretch}{1.12}
    \small
    \begin{tabular}{l l >{\raggedright\arraybackslash}p{7.8cm} c c}
        \toprule
        \textbf{Region} & \textbf{City} &
        \textbf{Collection Sites (Hashtags)} & \textbf{Type} & \textbf{Tier} \\
        \midrule
        \multirow{4}{*}{Southwest}
        & Chengdu
        & Chunxi Road; Taikoo~Li$^\dagger$; Global~Centre;
          IFS (International Finance~Square); Jiuyanqiao
        & O/E & 2 \\
        \cmidrule(l){2-5}
        & Chongqing   & Jiefangbei~CBD                            & O   & 2 \\
        \cmidrule(l){2-5}
        & Kunming     & Joy~City$^\dagger$; Mixc~Hub$^\dagger$    & E   & 3 \\
        \cmidrule(l){2-5}
        & Guiyang
        & Huaguoyuan Shopping~Centre; Wanda~Plaza$^\dagger$       & E   & 3 \\
        \midrule
        \multirow{7}{*}{East China}
        & Shanghai
        & Xintiandi; K11 Art~Mall; TX~Huaihai; Lujiazui~IFC       & O/E & 1 \\
        \cmidrule(l){2-5}
        & Hangzhou    & Hubin~Yintai; Binjiang~Mixc$^\dagger$; In77 & E  & 2 \\
        \cmidrule(l){2-5}
        & Nanjing
        & Deji~Plaza; Shuiyoucheng; Jingfeng Center; IFC          & M   & 2 \\
        \cmidrule(l){2-5}
        & Suzhou      & Yuanrong Times~Plaza; Suzhou~Center        & E   & 2 \\
        \cmidrule(l){2-5}
        & Ningbo      & Tianyi~Square; Ningbo~Mixc$^\dagger$       & E   & 2 \\
        \cmidrule(l){2-5}
        & Xiamen      & SM City~Plaza; Zhonghua~City               & E   & 3 \\
        \cmidrule(l){2-5}
        & Qingdao     & Qingdao~Mixc$^\dagger$; Hisense~Plaza      & E   & 2 \\
        \midrule
        \multirow{3}{*}{South China}
        & Guangzhou   & Taikoo~Hui$^\dagger$; Zhengjiahui; K11$^\dagger$ & E & 1 \\
        \cmidrule(l){2-5}
        & Shenzhen
        & Shenzhen~Mixc$^\dagger$; Yifang~City; COCO~Park         & E   & 1 \\
        \cmidrule(l){2-5}
        & Nanning
        & Nanning~Mixc$^\dagger$; Chaoyang~Square                 & E   & 3 \\
        \midrule
        \multirow{2}{*}{North China}
        & Beijing
        & SKP-S; Sanlitun Taikoo~Li$^\dagger$;
          Chaoyang Joy~City$^\dagger$; Xidan Joy~City$^\dagger$;
          Guomao~Mall; Wangjing~SOHO
        & E/O & 1 \\
        \cmidrule(l){2-5}
        & Tianjin
        & Tianjin Joy~City$^\dagger$; Hang~Lung~Plaza             & E   & 2 \\
        \midrule
        \multirow{2}{*}{Central China}
        & Changsha    & IFS; Wenheyou                             & E/O & 2 \\
        \cmidrule(l){2-5}
        & Wuhan
        & Tiandi~Block; Wushang~MALL; Wushang~Dream~Times         & M   & 2 \\
        \midrule
        \multirow{2}{*}{Northeast}
        & Shenyang
        & Zhongjie~Street; Shenyang~Mixc$^\dagger$                & O/E & 3 \\
        \cmidrule(l){2-5}
        & Harbin
        & Central~Avenue; Harbin Wanda~Plaza$^\dagger$            & O/E & 3 \\
        \midrule
        \multirow{2}{*}{Central Plains}
        & Zhengzhou   & David~City; Zhenghong~City                & E   & 2 \\
        \cmidrule(l){2-5}
        & Hefei       & Hefei~Mixc$^\dagger$; Yintai~Centre       & E   & 2 \\
        \midrule
        \multirow{2}{*}{Northwest}
        & Xi'an       & Datang Evernight~City                     & O   & 2 \\
        \cmidrule(l){2-5}
        & Urumqi
        & Urumqi Wanda~Plaza$^\dagger$; Youhao Department~Store   & E   & 3 \\
        \bottomrule
    \end{tabular}
    \vspace{3pt}
    \begin{minipage}{\linewidth}
        \footnotesize
        Note: City tiers follow the \textit{Chinese City Business Attractiveness
        Ranking}~\citep{peredy2024chinese}.
    \end{minipage}
\end{table*}

\section{Per-Class Segmentation Vocabulary}
\label{app:segvocab}

Table~\ref{tab:segvocab} lists the Grounding~DINO open-vocabulary prompt
terms used in the \ourname{} annotation pipeline.
Every \husic{} class (0--9) has its own set of semantically appropriate
object terms to maximise detection recall for class-specific content while
minimising false positives.
For Classes~0--5 (Spatially Relevant), Grounding~DINO detections are passed
to SAM~2 for instance mask generation, producing T3 pseudo-labels.
For Classes~6--9 (Non-Spatially Relevant), the same Grounding~DINO prompts
support the annotation and UGC filtering pipeline but \emph{do not} produce
SAM segmentation masks, consistent with the T3 evaluation scope stated in
Section~\ref{sec:task3}.

\begin{table*}[h]
	\centering
	\caption{Per-class Grounding~DINO prompt vocabulary for the \ourname{}
		annotation pipeline.}
	\label{tab:segvocab}
	\setlength{\tabcolsep}{3.5pt}
	\renewcommand{\arraystretch}{1.10}
	\small
	\begin{tabularx}{\textwidth}{c l >{\raggedright\arraybackslash}X}
		\toprule
		\textbf{ID} & \textbf{Class} & \textbf{Instance Segmentation Object Prompt Terms} \\
		\midrule
		\multicolumn{3}{l}{\textit{Spatially Relevant (Classes 0--5)}} \\[3pt]
		0 & Exterior urban spaces with people
		& person $\cdot$ crowd $\cdot$ pedestrian $\cdot$ building fa\c{c}ade
		$\cdot$ lawn $\cdot$ street lamp $\cdot$ glass curtain wall $\cdot$ sky
		$\cdot$ tree $\cdot$ shrub $\cdot$ fence $\cdot$ road $\cdot$ water
		$\cdot$ river $\cdot$ vehicle $\cdot$ sculpture $\cdot$ installation
		$\cdot$ pavement $\cdot$ street signage $\cdot$ fountain \\[3pt]
		1 & Exterior urban spaces without people
		& building fa\c{c}ade $\cdot$ glass curtain wall $\cdot$ wooden fa\c{c}ade
		$\cdot$ tree $\cdot$ shrub $\cdot$ lawn $\cdot$ sky $\cdot$ pavement
		$\cdot$ road $\cdot$ water $\cdot$ river $\cdot$ lantern $\cdot$ sculpture
		$\cdot$ installation $\cdot$ street lamp $\cdot$ signage $\cdot$ fence
		$\cdot$ bridge $\cdot$ water feature $\cdot$ fountain \\[3pt]
		2 & Interior urban spaces with people
		& person $\cdot$ shopper $\cdot$ crowd $\cdot$ retail shelf $\cdot$
		escalator $\cdot$ elevator $\cdot$ ceiling $\cdot$ floor tile $\cdot$
		glass partition $\cdot$ display case $\cdot$ door $\cdot$ indoor plant
		$\cdot$ wall $\cdot$ window $\cdot$ handrail $\cdot$ column \\[3pt]
		3 & Interior urban spaces without people
		& retail shelf $\cdot$ escalator $\cdot$ indoor corridor $\cdot$ ceiling
		$\cdot$ floor tile $\cdot$ marble floor $\cdot$ glass partition $\cdot$
		display case $\cdot$ wall $\cdot$ column $\cdot$ indoor plant $\cdot$
		elevator $\cdot$ door $\cdot$ window $\cdot$ lighting fixture
		$\cdot$ handrail \\[3pt]
		4 & Hotel or commercial lodging spaces
		& hotel bed $\cdot$ furniture $\cdot$ sofa $\cdot$ carpet $\cdot$
		marble floor $\cdot$ tile floor $\cdot$ wooden floor $\cdot$ ceiling
		$\cdot$ bathroom $\cdot$ window $\cdot$ curtain $\cdot$ lamp \\[3pt]
		5 & Private home interiors
		& sofa $\cdot$ bed $\cdot$ dining table $\cdot$ floor $\cdot$ ceiling
		$\cdot$ kitchen $\cdot$ bookshelf $\cdot$ wardrobe $\cdot$ window
		$\cdot$ lamp $\cdot$ carpet $\cdot$ wall \\[6pt]
		\midrule
		\multicolumn{3}{l}{\textit{Non-Spatially Relevant (Classes 6--9)}} \\[3pt]
		6 & Food or drink items
		& food dish $\cdot$ meal plate $\cdot$ dessert $\cdot$ beverage cup
		$\cdot$ coffee $\cdot$ drink bottle $\cdot$ bowl $\cdot$ chopsticks
		$\cdot$ spoon $\cdot$ dining table $\cdot$ person $\cdot$
		restaurant interior \\[3pt]
		7 & Retail products and merchandise
		& fashion clothing $\cdot$ shoes $\cdot$ cosmetics $\cdot$ product package
		$\cdot$ merchandise $\cdot$ retail shelf $\cdot$ bag $\cdot$ jewelry
		$\cdot$ electronics $\cdot$ store window $\cdot$ mannequin $\cdot$
		person \\[3pt]
		8 & Human-centered portrait
		& person $\cdot$ face $\cdot$ group photo $\cdot$ building fa\c{c}ade
		$\cdot$ sky $\cdot$ tree $\cdot$ floor $\cdot$ food $\cdot$ animal
		$\cdot$ vehicle $\cdot$ indoor background \\[3pt]
		9 & Other Non-spatial content
		& animal $\cdot$ person $\cdot$ vehicle $\cdot$ advertisement poster
		$\cdot$ text $\cdot$ QR code $\cdot$ screenshot $\cdot$ sculpture
		$\cdot$ meme $\cdot$ sky $\cdot$ plant $\cdot$ signage $\cdot$
		graphic design $\cdot$ logo $\cdot$ map $\cdot$ infographic $\cdot$
		chat record \\[3pt]
		\bottomrule
	\end{tabularx}
\end{table*}

\section{Ethical Considerations}
\label{app:ethics}

\textbf{Public content only.}
All collected posts originate from accounts whose visibility was explicitly
set to public by the account holder at the time of collection---the default
``open to all'' setting on Weibo.
No private, friends-only, or password-protected content was accessed.

\textbf{Platform compliance.}
Data collection followed Weibo's publicly documented terms of service for
academic research use.
The crawler respected rate limits and \texttt{robots.txt} directives, and no
authentication credentials of third parties were used.

\textbf{Research purpose.}
\ourname{} is designed exclusively for non-commercial academic research
into urban spatial perception and human behaviour in public commercial
environments, serving a clear public good by enabling evidence-based design
and planning guidance for more equitable and liveable urban spaces.
Dataset use is monitored; misuse may result in retraction.

\textbf{Privacy protection and data minimisation.}
The publicly distributed dataset omits original-resolution images, raw
usernames, and any metadata unnecessary for the three benchmark tasks.
Automated face detection, licence-plate recognition, and QR-code detection
were applied to all images, with all detected regions blurred; a manual
spot-check verified blurring completeness.
Images are released at a maximum side length of 512\,px, substantially
reducing re-identification risk.
The raw \textbf{4~TB} corpus is retained securely by the authors and will
not be publicly distributed.
The \texttt{Post~Text} field retains original Chinese to avoid translation
distortion (scientifically important for Task~2) but contains no directly
identifying information beyond what the original public post disclosed.

\textbf{Data-use agreement.}
Researchers wishing to access \ourname{} must agree to a data-use agreement
restricting use to non-commercial academic research and prohibiting
re-identification, surveillance, face recognition, account reconstruction,
and commercial profiling.

\clearpage
\newpage

\section*{NeurIPS Paper Checklist}

\begin{enumerate}

\item {\bf Claims}
    \item[] Question: Do the main claims made in the abstract and introduction accurately reflect the paper's contributions and scope?
    \item[] Answer: \answerYes{}
    \item[] Justification: The abstract and Section~\ref{sec:intro} clearly state the three contributions of the paper---the \ourname{} dataset, the \ourname{}-lib benchmarking library, and the \husic{} classification framework---and the experimental results in Sections~\ref{sec:exp_t1}, \ref{sec:t2_results}, and \ref{sec:exp_t3} directly substantiate each claimed contribution. Limitations and interpretive boundaries are articulated in Section~\ref{sec:discussion} and Appendix~\ref{app:discussion}.

\item {\bf Limitations}
    \item[] Question: Does the paper discuss the limitations of the work performed by the authors?
    \item[] Answer: \answerYes{}
    \item[] Justification: Limitations are discussed in Section~\ref{sec:discussion} (main body) and comprehensively enumerated in Appendix~\ref{app:discussion}, covering six specific limitations: class-agnostic T3 evaluation, geographic restriction to Chinese cities, class imbalance in the 2M corpus, incomplete LLaVA-1.5 100K training, T3 SAM oracle circularity, and the Chinese-language social media text domain gap.

\item {\bf Theory assumptions and proofs}
    \item[] Question: For each theoretical result, does the paper provide the full set of assumptions and a complete (and correct) proof?
    \item[] Answer: \answerNA{}
    \item[] Justification: The paper does not present theoretical results, lemmas, or proofs. All contributions are empirical (dataset, benchmarking library, experimental evaluation) and methodological (the \husic{} taxonomy derived from urban theory).

\item {\bf Experimental result reproducibility}
    \item[] Question: Does the paper fully disclose all the information needed to reproduce the main experimental results of the paper to the extent that it affects the main claims and/or conclusions of the paper (regardless of whether the code and data are provided or not)?
    \item[] Answer: \answerYes{}
    \item[] Justification: Full hyperparameter configurations for all three tasks are provided in Appendix~\ref{app:hyperparams}. Compute requirements are documented in Appendix~\ref{app:cost}. The corrected Task~2 multi-positive evaluation protocol is described in detail in Appendix~\ref{app:t2code}, including a quick-start guide for the evaluation script. Dataset splits, annotation procedures, and segmentation vocabulary are documented in Sections~\ref{sec:annotation} and~\ref{sec:task3} and Appendix~\ref{app:segvocab}. All data and code are publicly released (see footnote in abstract).

\item {\bf Open access to data and code}
    \item[] Question: Does the paper provide open access to the data and code, with sufficient instructions to faithfully reproduce the main experimental results, as described in supplemental material?
    \item[] Answer: \answerYes{}
    \item[] Justification: The dataset is hosted on Hugging Face (\texttt{huggingface.co/datasets/Yiwei-Ou/Urban-ImageNet}) and the benchmarking library \ourname{}-lib is released on GitHub (\texttt{github.com/yiasun/dataset-2}), as noted in the abstract footnote and Section~\ref{sec:benchmark}. The repository includes modular data loaders, fine-tuning pipelines, and the corrected multi-positive evaluation script described in Appendix~\ref{app:t2code}.

\item {\bf Experimental setting/details}
    \item[] Question: Does the paper specify all the training and test details (e.g., data splits, hyperparameters, how they were chosen, type of optimizer) necessary to understand the results?
    \item[] Answer: \answerYes{}
    \item[] Justification: Task-level evaluation protocols are described in Sections~\ref{sec:task1}, \ref{sec:task2}, and~\ref{sec:task3}. Full training details including optimizer, learning rate, batch size, training epochs, data augmentation, and hardware for all models and all three tasks are provided in Appendix~\ref{app:hyperparams}. Dataset split ratios (80:10:10) and multi-scale tier definitions are described in Section~\ref{sec:annotation}.

\item {\bf Experiment statistical significance}
    \item[] Question: Does the paper report error bars suitably and correctly defined or other appropriate information about the statistical significance of the experiments?
    \item[] Answer: \answerNo{}
    \item[] Justification: Error bars and confidence intervals are not reported in the main results tables. Task~1 results are obtained via five-fold cross-validation (noted in Sections~\ref{sec:task1} and~\ref{sec:exp_t1}), which provides implicit variance information, but variance estimates are not reported explicitly. Reporting error bars for the full model suite across three tasks and four dataset scales was computationally prohibitive; the five-fold cross-validation protocol is provided as a partial substitute for T1.

\item {\bf Experiments compute resources}
    \item[] Question: For each experiment, does the paper provide sufficient information on the computer resources (type of compute workers, memory, time of execution) needed to reproduce the experiments?
    \item[] Answer: \answerYes{}
    \item[] Justification: Appendix~\ref{app:cost} provides a detailed table listing hardware (CPU M2 Pro, A100-40G, A100-80G, H100-80G), training time in seconds, per-image inference time, and accuracy for each model--tier combination across all three dataset scales. The table also notes which experiments were not completed due to resource constraints (LLaVA-1.5 at 100K scale).
    
\item {\bf Code of ethics}
    \item[] Question: Does the research conducted in the paper conform, in every respect, with the NeurIPS Code of Ethics \url{https://neurips.cc/public/EthicsGuidelines}?
    \item[] Answer: \answerYes{}
    \item[] Justification: The research involves only publicly available social media content, complies with the platform's terms of service, and implements privacy-preserving measures including face blurring, username anonymisation, and resolution reduction. A dedicated ethical considerations section is provided in Appendix~\ref{app:ethics}, covering platform compliance, data minimisation, anonymisation procedures, and a data-use agreement restricting non-commercial academic use.

\item {\bf Broader impacts}
    \item[] Question: Does the paper discuss both potential positive societal impacts and negative societal impacts of the work performed?
    \item[] Answer: \answerYes{}
    \item[] Justification: Appendix~\ref{app:ethics} discusses the positive societal impact of enabling evidence-based urban design and planning for more equitable and liveable spaces. Potential negative impacts are addressed through the data-use agreement (prohibiting re-identification, surveillance, face recognition, account reconstruction, and commercial profiling) and through privacy-preserving data release practices described in Section~\ref{sec:processing} and Appendix~\ref{app:ethics}.
    
\item {\bf Safeguards}
    \item[] Question: Does the paper describe safeguards that have been put in place for responsible release of data or models that have a high risk for misuse (e.g., pre-trained language models, image generators, or scraped datasets)?
    \item[] Answer: \answerYes{}
    \item[] Justification: The dataset was scraped from a public social media platform and poses potential re-identification risks. Safeguards described in Section~\ref{sec:processing} and Appendix~\ref{app:ethics} include: automated face, licence-plate, and QR-code blurring with manual spot-check verification; username anonymisation via opaque numerical identifiers; image resolution reduction to a maximum 512\,px side length; withholding the raw 4\,TB corpus; and a data-use agreement restricting use to non-commercial academic research and prohibiting surveillance and re-identification.

\item {\bf Licenses for existing assets}
    \item[] Question: Are the creators or original owners of assets (e.g., code, data, models), used in the paper, properly credited and are the license and terms of use explicitly mentioned and properly respected?
    \item[] Answer: \answerYes{}
    \item[] Justification: All baseline models (ResNet, EfficientNet, ViT, DeiT, CLIP, BLIP, BLIP-2, LLaVA-1.5, Mask R-CNN, Cascade Mask R-CNN, SAM) and comparison datasets (Places365, SUN, MS-COCO, Cityscapes, Flickr30K, LAION-5B, ADE20K, LVIS, Mapillary Vistas, Place Pulse~2.0, MMS-VPR, UrbanFeel) are properly cited throughout the paper, primarily in Section~\ref{sec:related} and Tables~\ref{tab:comparison} and~\ref{tab:t1_results}--\ref{tab:t3_main}. Third-party annotation tools (Grounding DINO, SAM~2) are cited in Sections~\ref{sec:annotation} and~\ref{sec:task3}. Data collection complied with Weibo's terms of service for academic research use, as described in Appendix~\ref{app:ethics}.

\item {\bf New assets}
    \item[] Question: Are new assets introduced in the paper well documented and is the documentation provided alongside the assets?
    \item[] Answer: \answerYes{}
    \item[] Justification: The \ourname{} dataset is documented via a Croissant metadata file (including Responsible AI fields) submitted with the paper, and through structured documentation in Sections~\ref{sec:dataset}--\ref{sec:annotation} and Appendices~\ref{app:husic}, \ref{app:geo}, \ref{app:t2schema}, and~\ref{app:segvocab}. The \ourname{}-lib benchmarking library is documented in Section~\ref{sec:benchmark} and includes executable code on GitHub. The \husic{} taxonomy is formally defined in Section~\ref{sec:husic} and Appendix~\ref{app:husic}. Dataset splits, annotation schema, and file structure are described in Section~\ref{sec:annotation}.

\item {\bf Crowdsourcing and research with human subjects}
    \item[] Question: For crowdsourcing experiments and research with human subjects, does the paper include the full text of instructions given to participants and screenshots, if applicable, as well as details about compensation (if any)? 
    \item[] Answer: \answerYes{}
    \item[] Justification: Manual annotation was conducted by three trained researchers (not external crowd workers) following a standardised guideline described in Section~\ref{sec:annotation}. Inter-rater reliability was measured on a 3,000-image shared calibration subset, yielding Cohen's $\kappa = 0.87$. The annotation process, disagreement resolution procedure (majority vote and guideline revision), and calibration protocol are described in Section~\ref{sec:annotation}. The annotators are research personnel, and compensation details are governed by their institutional arrangements.

\item {\bf Institutional review board (IRB) approvals or equivalent for research with human subjects}
    \item[] Question: Does the paper describe potential risks incurred by study participants, whether such risks were disclosed to the subjects, and whether Institutional Review Board (IRB) approvals (or an equivalent approval/review based on the requirements of your country or institution) were obtained?
    \item[] Answer: \answerYes{}
    \item[] Justification: The dataset consists entirely of publicly available social media posts; no new primary data collection involving human participants was conducted beyond the internal annotation process performed by research personnel. The ethical framework governing data collection, privacy protection, and research purpose is described in Appendix~\ref{app:ethics}, and collection complied with platform terms of service and relevant institutional review requirements for secondary use of publicly available online data.

\item {\bf Declaration of LLM usage}
    \item[] Question: Does the paper describe the usage of LLMs if it is an important, original, or non-standard component of the core methods in this research? Note that if the LLM is used only for writing, editing, or formatting purposes and does \emph{not} impact the core methodology, scientific rigor, or originality of the research, declaration is not required.
    \item[] Answer: \answerYes{}
    \item[] Justification: Vision-language models (CLIP, BLIP, BLIP-2) and a large vision-language model (LLaVA-1.5) are used as core experimental baselines for Task~1 and Task~2, and their use as evaluation models is central to the benchmark contribution. These are declared and cited in Sections~\ref{sec:task1}, \ref{sec:task2}, and Appendices~\ref{app:hyperparams} and~\ref{app:scaling}. Grounding DINO, used for open-vocabulary detection in the pseudo-label annotation pipeline, is also cited in Sections~\ref{sec:annotation} and~\ref{sec:task3}. No LLMs were used solely for writing or formatting.

\end{enumerate}

\end{document}